\documentclass{article}

\PassOptionsToPackage{numbers,compress}{natbib}
\usepackage[main, final]{neurips_2026}

\usepackage[utf8]{inputenc} 
\usepackage[T1]{fontenc}    
\usepackage{hyperref}       
\usepackage{url}            
\usepackage{booktabs}       
\usepackage{amsfonts}       
\usepackage{nicefrac}       
\usepackage{microtype}      

\usepackage{amssymb}
\usepackage{amsmath}
\usepackage{multirow}
\usepackage{graphicx}
\usepackage{subcaption}
\usepackage[most]{tcolorbox}
\usepackage[table]{xcolor}

\usepackage{hyperref}
\usepackage{url}
\usepackage{booktabs}

\definecolor{linkred}{RGB}{180,40,40}
\definecolor{linkblue}{RGB}{25,85,165}

\hypersetup{
    colorlinks=true,
    linkcolor=linkred,
    citecolor=linkblue,
    urlcolor=linkblue
}

\definecolor{titlebgcolor}{RGB}{82,140,182}
\definecolor{textbgcolor}{RGB}{219,232,241}

\definecolor{tablegreen}{RGB}{46,125,50}

\newtcolorbox{infobox}[1][]{
    colframe=titlebgcolor!60,
    colback=textbgcolor!29,
    fontupper=\normalfont,
    #1
}

\definecolor{ObservationBlue}{HTML}{77AADD}

\title{Do More Modalities Always Help? A Geometric Perspective on Missing-Modality Robustness}

\author{%
  \textbf{Songyuan Sui}\textsuperscript{1}
  \quad
  \textbf{Zhen Tan}\textsuperscript{2}
  \quad
  \textbf{Mohan Zhang}\textsuperscript{3}
  \\
  \textbf{Rana Muhammad Shahroz Khan}\textsuperscript{3}
  \quad
  \textbf{Xia Hu}\textsuperscript{1}
  \quad
  \textbf{Tianlong Chen}\textsuperscript{3}
  \\[1mm]
  \textsuperscript{1}Rice University
  \qquad
  \textsuperscript{2}Stevens Institute of Technology
  \\
  \textsuperscript{3}University of North Carolina at Chapel Hill
  \\
  \texttt{Songyuan.Sui@rice.edu}
}

\begin{document}

\maketitle

\begin{abstract}
Missing modality remains a longstanding challenge in multimodal learning. Existing methods typically address this issue through modality recovery or adaptive strategies. However, they overlook models' internal cross-modal dependencies formed during multimodal training, which later impair robustness. We systematically characterize a counterintuitive deployment-time failure mode: models trained on full modalities can underperform unimodal models when one modality is missing at inference time. This pattern appears across diverse architectures, such as fusion models, CLIP-style two-tower models, and vision-language models. We show that such degradation is closely associated with learned cross-modal dependencies in the principal parameter subspaces. Multimodal training induces structured rotations of these subspaces, particularly in cross-modal interaction layers. These rotations are associated with reduced task-aligned margins and larger task-aware representation harm under missing-modality inputs. We propose \textbf{Geodesic Unlearning (GU)}, a lightweight parameter-editing method that leverages Grassmannian subspace geometry for structured subspace correction to improve missing-modality robustness. It rotates the principal input subspace toward a unimodal reference along a geodesic path. We prove that this correction minimizes the distance to the reference within a fixed subspace-distance budget. Experiments across architectures and datasets show that GU improves performance under missing-modality inference while preserving full-modality accuracy, outperforming strong missing-modality robustness baselines. These findings support a geometric view of deployment-time missing-modality degradation and suggest localized subspace editing as a practical route for robustness correction.
\end{abstract}

\addtocontents{toc}{\protect\setcounter{tocdepth}{-1}}

\section{Introduction}

Multimodal learning has achieved strong performance across language, vision, and healthcare domains \cite{zhang-etal-2023-learning-language, dong2025insight, soenksen2022integrated}. Large language models (LLMs) also support reasoning and interpretation
beyond plain text, such as vision, time series, and relational databases \cite{dong2025insight, 10.1145/3774905.3793127,  10.1609/aaai.v40i26.39319, sui-etal-2025-chain}. In jointly trained multimodal models, combining complementary signals typically improves performance over their unimodal counterparts \cite{huang2021makes}. However, this advantage may not transfer to deployment, where a modality can become unavailable due to sensor failure, incomplete data collection, privacy constraints, or system limitations \cite{wu2024deep}. Prior work on imbalanced multimodal learning shows that joint training may under-optimize modality-specific representations \cite{peng2022balanced}. We study a distinct deployment-time manifestation: when one modality is completely absent at inference, a model trained with full modalities can underperform one trained only on the available modality. This setting usually arises in heterogeneous deployment environments with varying device or sensor availability \cite{ma2021smil, reza2024robust, geraghty2025learning}.

Across diverse architectures, we observe a recurring ordering where full-modality inference performs best, unimodal training performs next, and full-modality training with a missing modality performs worst. This suggests that missing-modality degradation is not merely due to absent input information, but is also linked to internal cross-modal dependencies shaped by multimodal training.

We characterize this degradation geometrically. At the parameter level, multimodal models exhibit structured principal-subspace rotations, measured by principal-angle deviations between matched multimodal and unimodal input subspaces, especially in cross-modal interaction layers. Rotation magnitude is strongly associated with the missing-modality gap. At the representation level, misclassified samples exhibit larger task-aware harm (negative signed task margin) than correctly classified samples at intermediate multimodal interaction stages under missing-modality inference. This gap persists toward the classifier. These localized parameter- and representation-space signatures motivate targeted parameter-level correction beyond recovery-based or adaptive methods.

Motivated by these findings, we formulate missing-modality robustness as a special form of \emph{machine unlearning}: attenuating harmful cross-modal dependencies under missing inputs while preserving useful multimodal functionality. We propose \textbf{Geodesic Unlearning (GU)}, a lightweight post-training editing method. Given a multimodal model and a matched unimodal reference, GU edits \textbf{one target layer} selected by an angle--harm criterion. It treats the selected principal subspace as a point on the \emph{Grassmannian}, rotates its input-side basis toward the unimodal reference along a geodesic path, and reconstructs the layer while retaining the principal coefficients, output factor, and residual. Our theoretical analysis establishes the geometric optimality of this correction under a fixed subspace-distance budget and shows that the path advances all principal angles proportionally.

Experiments across fusion models, CLIP-style two-tower architectures, vision-language models (VLMs), and multiple datasets further show that GU improves performance on missing-modality tasks while largely preserving full-modality accuracy. Comparisons with multimodal-specific methods, general training and adaptation baselines, and parameter-editing controls support GU's favorable robustness--preservation trade-off, with the strongest gains when harmful dependencies are localized in editable interaction layers.

In summary, our contributions are as follows:
\begin{itemize}
    \item We systematically characterize deployment-time degradation when one modality is entirely unavailable at inference across three major multimodal architectures.
    \item We provide a geometric account of this degradation, supported by parameter- and representation-level evidence, showing that it is strongly associated with principal subspace rotations and task-aware representational harm in cross-modal interaction layers.
    \item We propose Geodesic Unlearning, a lightweight, theoretically grounded Grassmannian subspace editing method that applies across models to improve missing-modality robustness while preserving full-modality utility.
\end{itemize}

\section{Related work}

\noindent\textbf{Missing Modality} remains a challenge in multimodal learning \cite{sebe2005multimodal, zhang2023distilling, wu2024deep}. Existing approaches address this challenge by recovering unavailable information or improving how models use available modalities. Recovery-based methods impute missing inputs at the data level \cite{yang2024incomplete, hussein2024sensor, sun2024similar, wang2023incomplete, zhang2024unified} or representation level \cite{ma2021smil, wang2023learnable, zheng2024learning, sun2024redcore, gao2025deep, wang2025iterative}, relying on cross-modal correlations and reconstruction quality. Adaptive methods accommodate missing inputs through flexible architectures \cite{xue2023dynamic, suris2023vipergpt, yun2024flex}, parameter adaptation \cite{hu2021loralowrankadaptationlarge, zhao2026mora}, or training--inference strategies \cite{wang2023multi, yao2024drfuse, hu2024deep, xu2025distilled, wen2026simo, du2026inferencetime}. Complementary modality-balancing methods, including OGM-GE \cite{peng2022balanced}, AGM \cite{Li_2023_ICCV}, DnR \cite{wei2024diagnosing}, and MCR \cite{NEURIPS2025_f5784de2}, regulate the relative learning contributions of modalities during joint training. Our work focuses on harmful parameter dependencies retained when a modality is completely absent at deployment. We diagnose their task-relevant principal-subspace geometry and correct them in an already-trained model, providing a model-side approach that can complement training-time robustness objectives.

\noindent\textbf{Geometric Understanding of Neural Networks} has been studied from both representation and parameter perspectives. At the representation level, prior work reveals structured feature geometry, including simplex class structure \cite{papyan2020prevalence}, representational similarity across models \cite{kornblith2019similarity}, and margin-based geometry related to generalization and robustness \cite{elsayed2018large, lyu2022improving}. At the parameter level, recent studies show that model parameters and gradients often concentrate in low-dimensional subspaces, motivating low-rank adaptation and subspace-based optimization \cite{hu2021loralowrankadaptationlarge, zhao2024galore, robert2025ldadam}. More recently, Grassmannian methods model the evolution of such subspaces and enable structure-preserving updates through geodesic paths \cite{edelman1998geometry, he2011online, NEURIPS2025_2d62cb71}. Our work connects these geometric perspectives to missing-modality robustness by showing that multimodal training induces structured rotations in principal parameter subspaces, which are associated with task-aware representation harm under missing inputs.

\noindent\textbf{Machine Unlearning (MU)} was originally proposed in the privacy domain to remove the influence of specific training data without full retraining \cite{7163042, bourtoule2021machine}. Existing methods span data-side interventions \cite{zhang2023unlearnable, sun2024unseg}, training-time updates \cite{chen2025score, NEURIPS2025_1f6af963}, decoding-time control \cite{golatkar2024cpr, prashant2025tokenswap}, and training-free parameter editing \cite{ilharco2023editing, biswas2025cure}. Beyond privacy, MU has been explored to improve model behavior in scenarios such as continual learning \cite{wang2024a}, domain adaptation \cite{basak2024forget}, and domain generalization \cite{zhao2026unlearning}. Motivated by our finding that multimodal training induces harmful cross-modal dependencies under missing inputs, we repurpose MU to selectively attenuate these dependencies while preserving useful multimodal capability. To the best of our knowledge, our work is the first to formulate missing-modality robustness as an unlearning problem and address it through geometric parameter editing.

\begin{figure}
    \centering
    \includegraphics[width=1.0\linewidth]{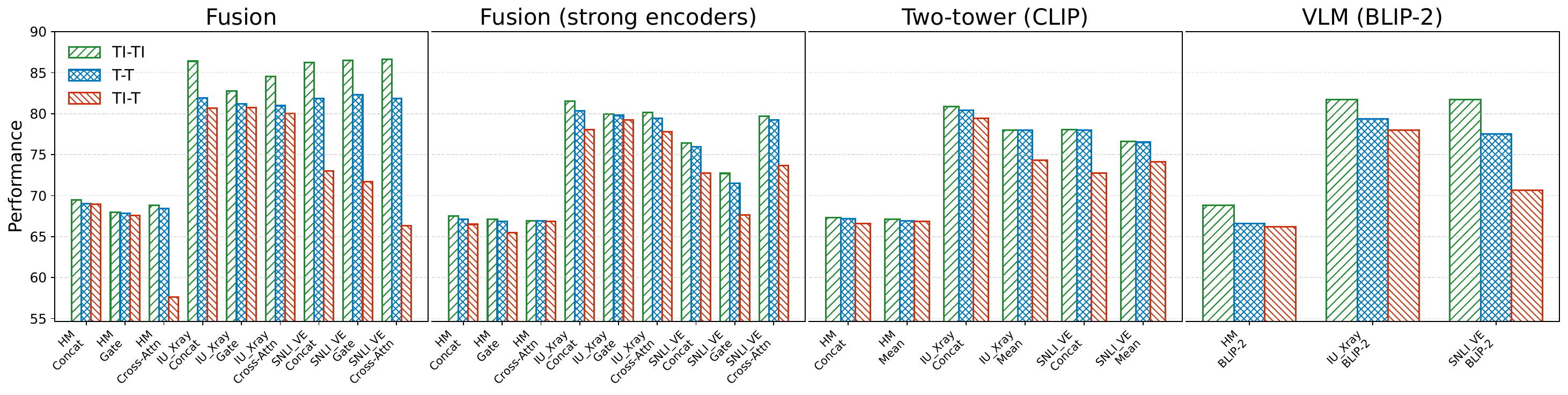}
    \caption{Missing-modality degradation across architectures and datasets. F1 scores are reported for TI-TI, T-T, and TI-T settings. In 26 of the 27 settings, we observe $\mathrm{TI\text{-}TI} > \mathrm{T\text{-}T} > \mathrm{TI\text{-}T}$, revealing a recurring deployment-time missing-modality degradation pattern.}
    \label{fig:analysis_degradation}
\end{figure}

\section{Problem setup and preliminaries}

\subsection{Deployment-time missing modality setting}

We consider a multimodal learning problem over modality set $\mathcal{M}=\{1,\dots,M\}$, where each modality $m\in\mathcal{M}$ has input space $\mathcal{X}_m$. The full-modality input space is $\mathcal{X}_{\mathrm{full}}=\prod_{m\in\mathcal{M}}\mathcal{X}_m$, and a full input is denoted by $x=(x_1,\dots,x_M)$. We study the \textbf{deployment-time} missing modality setting, where models are trained on full-modality inputs but evaluated with one or more modalities unavailable at inference. We introduce a null symbol $\perp$ and define $\mathcal{X}_{\mathrm{miss}} = \prod_{m\in\mathcal{M}}(\mathcal{X}_m\cup\{\perp\})$. If modality $m$ is missing at deployment, the corresponding input is $x_{-m} = (x_1,\dots,x_{m-1},\perp,x_{m+1},\dots,x_M)$. In practice, $\perp$ can be implemented as a masking token, zeroed feature stream, or learned null embedding. We focus on the practically important \textbf{single missing modality} case. For bimodal systems such as text-image models, the same modality is \textbf{entirely} absent for every test example in the missing-modality evaluation, while the other remains available.

\subsection{Evaluation protocol and missing-modality gap}
\label{sec:pre_eval}

We instantiate the setting with bimodal text-image models, where \textbf{T} denotes text and \textbf{I} denotes image. We consider three training--testing regimes: \textbf{TI-TI} trains and tests with text+image; \textbf{T-T} trains and tests with text only; and \textbf{TI-T} trains with text+image but tests with text only. Unless otherwise stated, we study missing-image evaluation, where text remains available, and the image is removed at inference. We define the \textbf{missing-modality gap} as $\Delta_{\mathrm{miss}} = \mathrm{Perf}(\text{TI-TI}) - \mathrm{Perf}(\text{TI-T})$ and the \textbf{unimodal robustness gap} as $\Delta_{\mathrm{uni}} = \mathrm{Perf}(\text{T-T}) - \mathrm{Perf}(\text{TI-T})$, where $\mathrm{Perf}(\cdot)$ denotes the task metric, instantiated as F1 scores in our main experiments. The former measures performance loss after removing the image modality at deployment, while the latter measures the extent to which the missing-modality multimodal model underperforms a text-specialized model.

\subsection{Parameter subspace geometry}
\label{sec:pre_para_subspace}

To study how multimodal training changes layer parameters, we analyze the principal singular subspaces of selected weight matrices. For a layer weight matrix $W \in \mathbb{R}^{d_o \times d_i}$ with SVD $W = U \Sigma V^\top$, we use the top-$k$ singular vectors $U_k=[u_1,\dots,u_k]$ and $V_k=[v_1,\dots,v_k]$ to define the principal output and input subspaces, respectively. The right singular vectors $V_k$ identify the dominant input directions the layer reads, while $U_k$ specifies the associated output directions. Since $V_k$ and $V_kQ$ span the same subspace for any orthogonal $Q\in\mathbb{R}^{k\times k}$, the basis-invariant object is $S=\operatorname{span}(V_k)\in\mathrm{Gr}(k,d_i)$; the output subspace is defined analogously on $\mathrm{Gr}(k,d_o)$. Given a multimodal model $W_{\mathrm{TI}}$ and a unimodal reference model $W_{\mathrm{T}}$, we compare their corresponding input and output subspaces to quantify how multimodal training changes parameter geometry, using standard metrics such as principal angles, subspace overlap, and Frobenius distances.

\begin{table}[t]
\caption{Representative model families used in our analysis. We cover distinct multimodal integration mechanisms, including explicit fusion architectures, aligned two-tower encoders, and VLMs.}
\label{tab:analysis_models}
\centering
\begin{tabular}{lll}
\toprule
Model Family & Backbone(s) & Variants \\
\midrule
Fusion &
RoBERTa \cite{liu2019roberta} + ResNet50 \cite{koonce2021resnet} &
Concat, Gate, Cross-Attention \\

Fusion (strong encoders) &
Qwen \cite{qwen2025qwen25technicalreport} + OpenCLIP \cite{ilharco2021openclip} &
Concat, Gate, Cross-Attention \\

Two-tower &
OpenCLIP &
Mean Pooling, Late Concat \\

VLM &
BLIP-2 \cite{li2023blip} &
Classifier Head \\
\bottomrule
\end{tabular}
\end{table}

\section{Empirical analysis of missing-modality degradation}
\label{sec:analysis}

\subsection{Analysis scope: architectures and benchmarks}
\label{sec:analysis_scope}

As summarized in Table~\ref{tab:analysis_models}, our analysis covers three representative families of multimodal architectures: fusion models, two-tower dual-encoder models, and vision-language models. This partition follows recent surveys that categorize multimodal systems by architectural organization and fusion strategy \cite{zhao2024deep, han2025multimodal, li2026multimodal}. The selected systems differ in how they encode, align, and fuse modalities. It allows us to examine whether deployment-time missing-modality degradation is tied to a specific architecture or appears more broadly. All analyses are conducted under the deployment-time missing-image setting defined in Section~\ref{sec:pre_eval}. Dataset details, evaluation metrics, and implementation protocols are provided in Section~\ref{sec:experiments}.

\subsection{RQ1: Is missing-modality degradation consistent across architectures?}
\label{sec:analysis_missing_failure}

\begin{infobox}
\phantomsection\label{obs:1}
\textbf{Observation 1.}
Across architectures and datasets, full-modality inference performs best, while missing-modality inference generally underperforms unimodal training.
\end{infobox}

Prior work on imbalanced multimodal learning has shown that joint multimodal training may under-optimize modality-specific representations~\citep{peng2022balanced}. We examine a distinct deployment-time question: whether a model trained with full modalities remains robust when one modality is entirely unavailable at inference. Figure~\ref{fig:analysis_degradation} compares the three regimes in Section~\ref{sec:pre_eval} across 27 architecture--dataset combinations from Table~\ref{tab:analysis_models}.
The ordering
$\mathrm{Perf}(\text{TI-TI}) >
\mathrm{Perf}(\text{T-T}) >
\mathrm{Perf}(\text{TI-T})$
holds in 26 of 27 settings, while $\mathrm{Perf}(\text{T-T}) > \mathrm{Perf}(\text{TI-T})$ holds in all 27 settings. The complete numerical results are provided in Appendix~\ref{app:missing_modality_degradation}.

Across the 27 settings, the missing-modality gap averages 4.74 F1 points but ranges from 0.08 to 20.29. Generally, architectures with more explicit cross-modal interaction, such as cross-attention fusion, tend to exhibit larger missing-modality gaps than those with weaker cross-modal coupling, such as gated fusion or mean-pooled dual encoders. This suggests that degradation is not only a consequence of removing input information, but is also associated with internal cross-modal dependencies formed during multimodal training. These dependencies may be analyzed from a parameter-subspace perspective.

\begin{figure}[t]
\centering

\begin{subfigure}{0.46\linewidth}
    \centering
    \includegraphics[width=\linewidth]{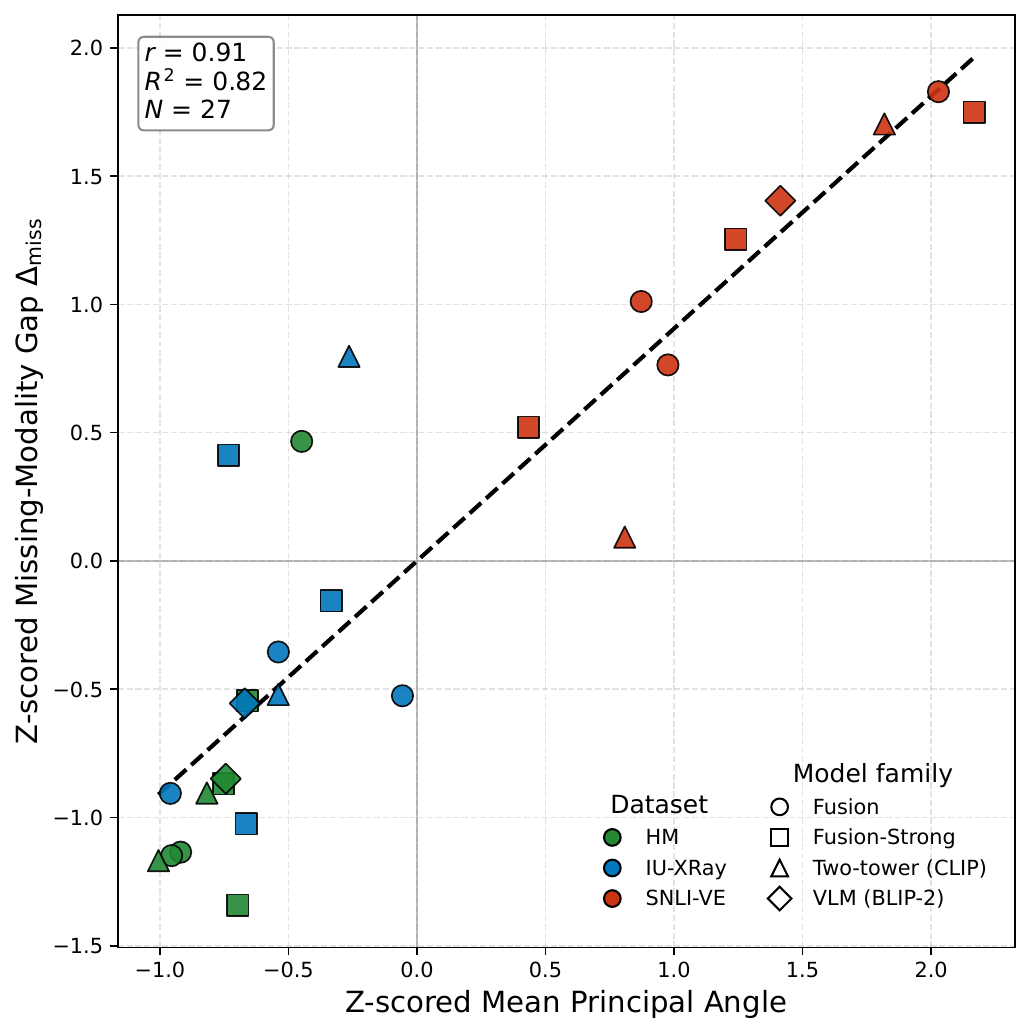}
    \caption{Normalized correlation across model families.}
\end{subfigure}
\hfill
\begin{subfigure}{0.48\linewidth}
    \centering
    \includegraphics[width=\linewidth]{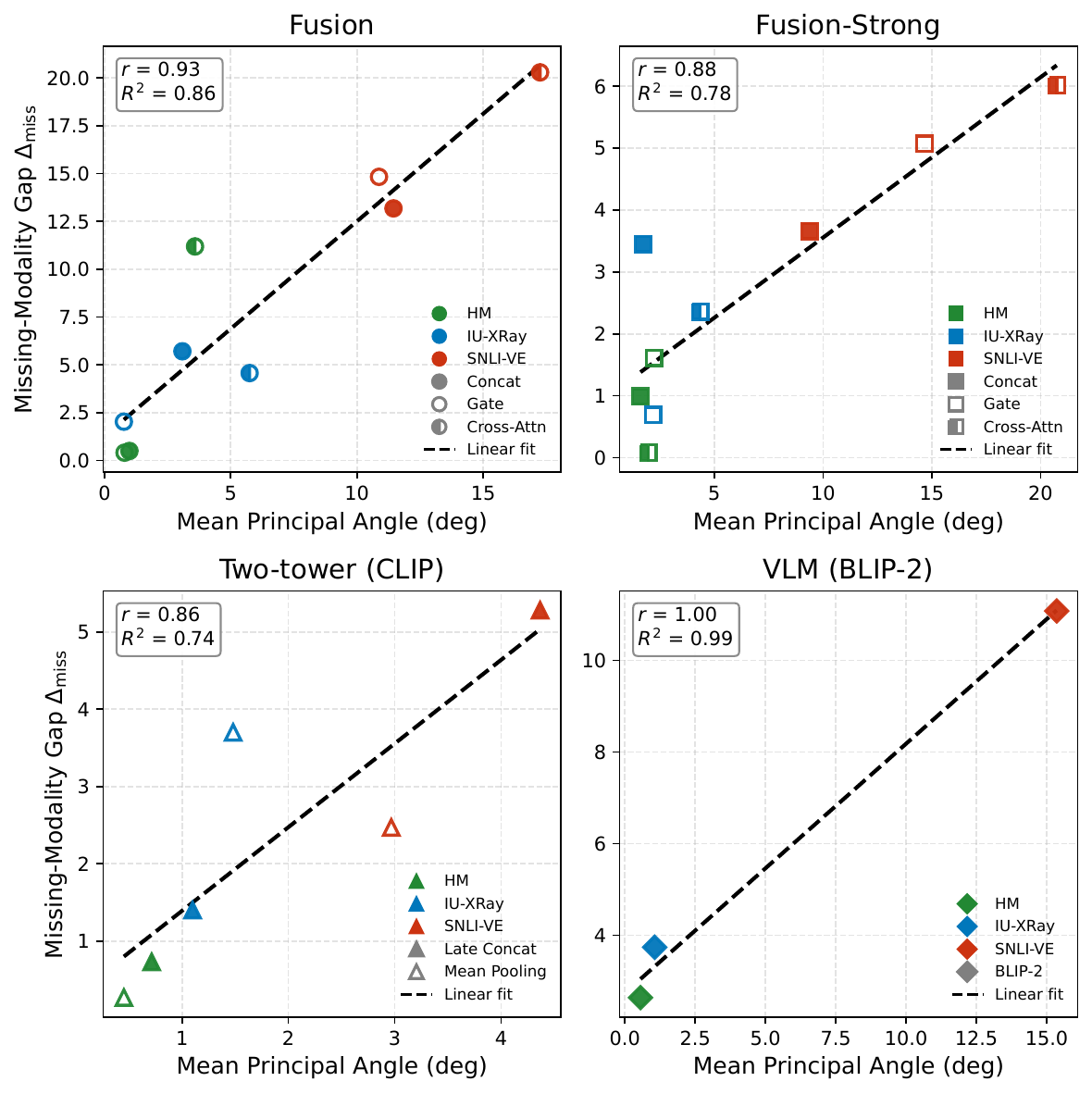}
    \caption{Raw correlation within each architecture.}
\end{subfigure}

\caption{Principal subspace rotation correlates with missing-modality degradation.
(a) Within-family z-score normalization reveals a global positive association.
(b) Raw within-family measurements show consistent positive associations without normalization.}
\label{fig:analysis_angle_gap}
\end{figure}

\subsection{RQ2: Does parameter geometry track degradation?}
\label{sec:analysis_para_subspace}

\begin{infobox}
\phantomsection\label{obs:2}
\textbf{Observation 2.}
Larger principal subspace rotations in cross-modal interaction layers are strongly and positively associated with larger missing-modality gaps.
\end{infobox}

We examine whether the degradation in Section~\ref{sec:analysis_missing_failure} has a measurable parameter-space signature. For each candidate layer, we compare the TI-trained model with its T-only reference using the principal singular subspaces defined in Section~\ref{sec:pre_para_subspace}. We measure their divergence by the mean principal angle $\bar{\theta}=k^{-1}\sum_{i=1}^{k}\theta_i$, where $\theta_i=\arccos\sigma_i\!\left(V_{\mathrm{TI},k}^{\top}V_{\mathrm{T},k}\right)$ and $\sigma_i(\cdot)$ denotes the $i$th singular value.

Figure~\ref{fig:analysis_angle_gap} shows that interaction-layer subspace rotation strongly tracks missing-modality degradation. After applying within-family z-score normalization to account for scale differences, Figure~\ref{fig:analysis_angle_gap}(a) shows a strong positive association between mean principal angle and missing-modality gap over all 27 architecture--dataset settings ($r=0.91$, $R^2=0.82$). The same relationship also appears in raw within-family measurements, as shown in Figure~\ref{fig:analysis_angle_gap}(b). It includes fusion models ($r=0.93$, $R^2=0.86$), stronger-encoder fusion models ($r=0.88$, $R^2=0.78$), and CLIP-style two-tower models ($r=0.86$, $R^2=0.74$). The BLIP-2 trend is also positive, but should be interpreted descriptively because it contains only three points. On SNLI-VE, Fusion Cross-Attention exhibits a gap of $20.29$ F1 points, with mean input/output principal angles of $16.00^\circ/16.57^\circ$ at \texttt{attn\_v\_from\_t\_in\_proj}, versus $1.95^\circ/0.00^\circ$ at the final classifier (Appendix~\ref{app:subspace_rotation}).

These results indicate a strong positive association that is not driven solely by normalization. They also suggest that multimodal training induces nonuniform parameter shifts, with the most informative geometric changes concentrated in cross-modal interaction layers. Principal-angle deviation therefore provides a geometric signature of missing-modality failure. The strength and consistency of this association motivate interaction-layer subspaces as localized targets for robustness correction.

\subsection{RQ3: Where does representation harm concentrate?}
\label{sec:analysis_repr_harm}

\begin{infobox}
\phantomsection\label{obs:3}
\textbf{Observation 3.}
Final-error versus final-correct task-aware harm is consistently represented at intermediate stages of multimodal interaction and often persists near the classifier.
\end{infobox}

\begin{figure}
    \centering
    \includegraphics[width=0.95\linewidth]{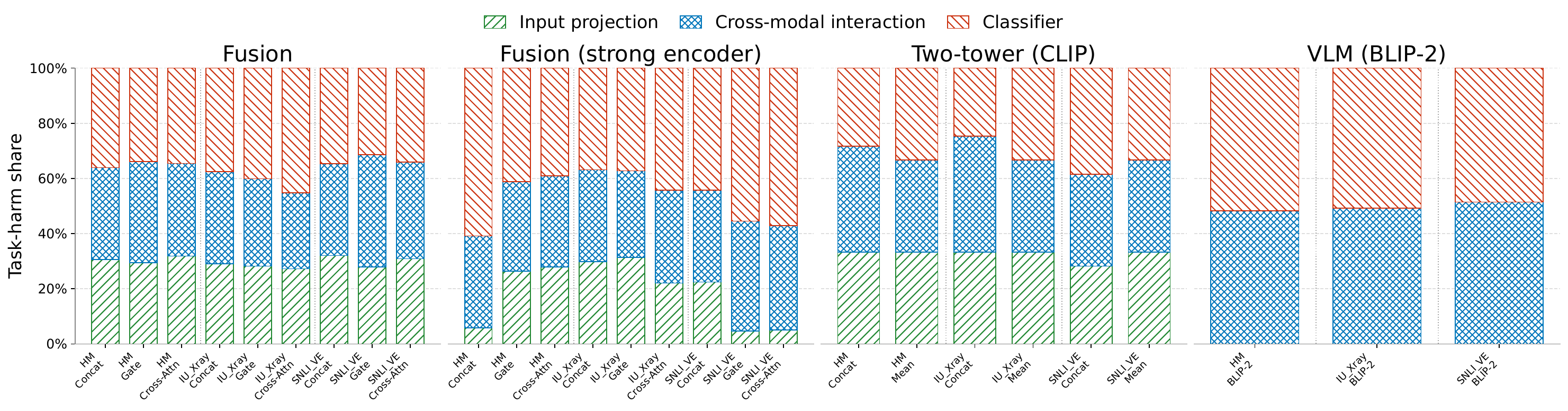}
    \caption{Stage-wise distribution of task-aware representation harm. The final-error versus final-correct harm gap appears at cross-modal interaction stages and often persists near the classifier.}
    \label{fig:analysis_repr_harm}
\end{figure}

We next localize task-relevant representation shifts induced by missing modality inputs.
For our binary classification tasks, we use a centroid-margin diagnostic at each probed stage.
Let $\mu_{+}$ and $\mu_{-}$ be the class centroids computed from representations at that stage, with task direction $d=\mu_{+}-\mu_{-}$ and midpoint $c=(\mu_{+}+\mu_{-})/2$.
For representation $h(x)$ and label $y\in\{+1,-1\}$, we define the signed task margin as $m(x)=y\cdot(h(x)-c)^\top \hat d$, where $\hat d=d/\|d\|_2$, and define $\mathrm{Harm}(x)=-m(x)$.
The error and correct sets, $\mathcal{E}$ and $\mathcal{C}$, are defined by the final prediction under missing-modality evaluation. We then probe intermediate representations using the same partition and compute
$\Delta_{\mathrm{harm}}=\mathbb{E}_{x\in\mathcal{E}}[\mathrm{Harm}(x)]-\mathbb{E}_{x\in\mathcal{C}}[\mathrm{Harm}(x)]$.

Figure~\ref{fig:analysis_repr_harm} summarizes the stage-wise distribution of the final-error versus final-correct harm gap. Bars are normalized within each dataset and model setting, so they indicate where the gap concentrates rather than its absolute magnitude. Final-error samples exhibit larger harm near the classifier, but the gap is already evident at cross-modal interaction stages before the final decision layer. These stages account for a substantial share of the gap across settings, consistent with Section~\ref{sec:analysis_para_subspace}, where the strongest subspace rotations also appear in cross-modal interaction layers. Classifier-stage harm in several settings is therefore better interpreted as downstream persistence than as a classifier-only effect. Detailed layer-wise results are provided in Appendix~\ref{app:repr_harm}.

Together, Sections~\ref{sec:analysis_missing_failure}--\ref{sec:analysis_repr_harm} provide the empirical basis for our method design: degradation is broadly observed, strongly associated with interaction-layer subspace rotations, and accompanied by task-aware representation shifts concentrated around multimodal interaction stages. We therefore use the identified interaction-layer subspace as a localized target for robustness correction.

\section{Geodesic Unlearning}

\subsection{Overview}
\label{sec:method_overview}

Section~\ref{sec:analysis} identifies cross-modal interaction layers as localized targets. Their principal subspace rotations track missing-modality degradation, and their representations exhibit task-aware harm. Given a trained multimodal model and a matched unimodal reference, we propose \textbf{Geodesic Unlearning (GU)} to perform a lightweight post-training correction of one selected layer. GU selects the target layer using an angle-harm criterion, rotates its principal input-side subspace toward the corresponding unimodal subspace along a Grassmannian geodesic, and reconstructs the layer while retaining the principal coefficients, output factor, and residual structure. This selectively attenuates harmful cross-modal dependencies under missing inputs while preserving useful multimodal functionality. The same subspace-editing procedure applies to fusion, CLIP-style two-tower, and BLIP-2 models, with only the target layer varying by architecture, enabling a unified post-training correction.

\subsection{Target-layer selection}
\label{sec:method_layer_selection}

For each model, GU selects one target layer from the interaction and nearby projection/classifier layers analyzed in Section~\ref{sec:analysis}. For each candidate layer $\ell$, let $\theta_\ell$ denote the mean principal angle between the top-$k$ input subspaces of the TI model and T reference, and let $\Delta_{\mathrm{harm},\ell}$ denote the final-error versus final-correct harm gap measured at layer $\ell$ under missing-modality evaluation.
We rank layers on the development set using
\begin{equation}
s_\ell = z(\theta_\ell) + z(\Delta_{\mathrm{harm},\ell}),
\label{eq:target_layer_score}
\end{equation}
where $z(\cdot)$ denotes z-score normalization across candidate layers within the same model. The layer with the largest $s_\ell$ is edited; the test set is reserved only for final evaluation.

\subsection{Subspace realignment and reconstruction}
\label{sec:method_realignment_reconstruction}

Given the selected multimodal layer weight $W_{\mathrm{TI}}\in\mathbb{R}^{d_o\times d_i}$ with SVD $W_{\mathrm{TI}}=U_{\mathrm{TI}}\Sigma_{\mathrm{TI}}V_{\mathrm{TI}}^\top$, we decompose it into a top-$k$ principal component and a residual:
\begin{equation}
W_{\mathrm{TI}}
=
U_{\mathrm{TI},k}\Sigma_{\mathrm{TI},k}V_{\mathrm{TI},k}^{\top}
+
R_{\mathrm{TI}},
\qquad
R_{\mathrm{TI}}
=
W_{\mathrm{TI}}
-
U_{\mathrm{TI},k}\Sigma_{\mathrm{TI},k}V_{\mathrm{TI},k}^{\top}.
\label{eq:gu_decomposition}
\end{equation}
Our diagnostics localize task-relevant deviation in the principal input subspace, so GU edits this orientation. Let $V_{\mathrm{T},k}$ be an orthonormal basis of the corresponding top-$k$ input subspace of the matched unimodal reference.
GU computes an orthonormal basis for the edited subspace
\begin{equation}
V(\eta)=\mathrm{Geodesic}(V_{\mathrm{TI},k},V_{\mathrm{T},k};\eta),
\qquad \eta\in[0,1],
\label{eq:gu_geodesic_main}
\end{equation}
where $\eta$ controls the edit strength: $\eta=0$ leaves the multimodal subspace unchanged, while $\eta=1$ reaches the unimodal-reference endpoint along the geodesic path.
The explicit geodesic construction is provided in Appendix~\ref{app:gu_geodesic_details}. The edited parameter is reconstructed as
\begin{equation}
W'
=
R_{\mathrm{TI}}
+
U_{\mathrm{TI},k}\Sigma_{\mathrm{TI},k}V(\eta)^\top .
\label{eq:gu_reconstruction}
\end{equation}
The reconstruction retains the TI output factor $U_{\mathrm{TI},k}$, principal coefficients $\Sigma_{\mathrm{TI},k}$, and residual $R_{\mathrm{TI}}$, whereas direct weight interpolation also changes these components.

\noindent\textbf{Geometric rationale.}
Let $S_{\mathrm{TI}}=\operatorname{span}(V_{\mathrm{TI},k})$, $S_{\mathrm T}=\operatorname{span}(V_{\mathrm T,k})$, and $S_\eta=\operatorname{span}(V(\eta))$. Write $d$ for the Grassmann geodesic distance, the $\ell_2$ norm of the principal-angle vector, and $D=d(S_{\mathrm{TI}},S_{\mathrm T})$. For any candidate subspace $S$ satisfying $d(S_{\mathrm{TI}},S)\leq\eta D$, the minimizing geodesic gives
\begin{equation}
d(S_{\mathrm{TI}},S_\eta)=\eta D,
\qquad
d(S_\eta,S_{\mathrm T})=(1-\eta)D\leq d(S,S_{\mathrm T}).
\label{eq:gu_edit_budget}
\end{equation}
Thus, GU reaches a subspace as close to the reference as any edit within the same subspace-distance budget. It advances through an $\eta$ fraction of each principal angle, whereas linear interpolation of aligned bases followed by orthonormalization generally yields angle-dependent progress. Appendix~\ref{app:gu_geometric_rationale} provides the proof and the linear-path comparison.
Controlled comparisons of target-layer selection and editing strategies are reported in Appendix~\ref{app:controlled_interventions}.

\section{Experiments}
\label{sec:experiments}

\subsection{Experimental setup}

\noindent\textbf{Datasets and metrics.}
We evaluate GU on three multimodal benchmarks: \textit{Hateful Memes} \cite{kiela2020hateful} for social multimodal classification, \textit{SNLI-VE} \cite{xie2019visual} for visual entailment, and \textit{IU-XRay} \cite{demner2016preparing} for medical vision-language prediction. These three benchmarks are formulated as binary classification tasks under the deployment-time missing-modality setting. We report F1 as the primary metric to account for potential class imbalance. More dataset information is provided in Appendix~\ref{app:dataset_details}.
We further evaluate GU on \textit{A-OKVQA} \cite{schwenk2022okvqa}, a multiple-choice VQA benchmark, in Appendix~\ref{app:aokvqa}.

\noindent\textbf{Model families.}
We use the model families summarized in Table~\ref{tab:analysis_models}. For each dataset and model family, we train a full multimodal model with text and image inputs, denoted as the TI model. We train a matched T reference using the same overall parameterization, where the image branch is replaced by a null input during both training and inference. This implements the T-T regime while keeping layer shapes aligned for subspace comparison and editing.

\noindent\textbf{Baselines.}
Our comparisons cover three groups.
\textbf{Multimodal-specific methods} include SMIL \cite{ma2021smil}, Flex-MoE \cite{yun2024flex}, and DyMo \cite{du2026inferencetime} in a unified fusion setting. We further evaluate MoRA \cite{zhao2026mora} and OGM-GE \cite{peng2022balanced} on compatible architectures beyond this setting.
\textbf{General training and adaptation baselines} include post-hoc LoRA \cite{hu2021loralowrankadaptationlarge} and Dual CE. Dual CE jointly supervises full-input and missing-image paths.
\textbf{Parameter-editing controls} include layer interpolation (Weight), following parameter-space editing ideas \cite{ilharco2023editing}, and aligned-subspace linear interpolation (Linear). These controls isolate the effects of the edited object and interpolation path.
Appendix~\ref{app:baselines} details their definitions, architecture coverage, and protocols. Additional comparisons are reported in Appendices~\ref{app:additional_baselines} and~\ref{app:controlled_interventions}.

\noindent\textbf{Implementation.}
GU is applied after training. For the main comparison with missing-modality baselines, we use a unified RoBERTa--ResNet50 cross-attention fusion setting. For cross-family robustness and preservation evaluation, we apply GU and layer interpolation to all model families in Table~\ref{tab:analysis_models}. We evaluate missing-image recovery alongside full-modality preservation. Baseline protocols and implementation details are provided in Appendices~\ref{app:baselines} and~\ref{app:implementation_details}.

\subsection{Main results in the deployment-time missing-modality setting}

\begin{table}[t]
\caption{Main missing-image performance (TI-T, F1 $\uparrow$). Values are three-run means. \underline{Underline} and \textbf{bold} denote the second-best and best results, and parentheses show gains over the second best.}
\label{tab:exp_main_performance}
\centering

\begin{tabular}{lccccc>{\columncolor{cyan!10}}c}
\toprule
Dataset
& Null-token
& SMIL
& Flex-MoE
& DyMo
& Interpolation
& \textbf{GU (Ours)} \\
\midrule

HM
& 57.63
& 56.60
& 47.58
& 48.82
& \underline{58.50}
& $\mathbf{60.88}$
  \hspace{0.3em}{\textcolor{green!60!black}{(+2.38)}} \\

IU-XRay
& 80.00
& 79.21
& 71.02
& 79.19
& \underline{80.14}
& $\mathbf{81.41}$
  \hspace{0.3em}{\textcolor{green!60!black}{(+1.27)}} \\

SNLI-VE
& 66.35
& \underline{72.20}
& 70.74
& 61.30
& 67.29
& $\mathbf{74.41}$
  \hspace{0.3em}{\textcolor{green!60!black}{(+2.21)}} \\

\bottomrule
\end{tabular}
\end{table}

\begin{figure}[t]
\centering
\begin{subfigure}{0.44\linewidth}
    \centering
    \includegraphics[width=\linewidth]{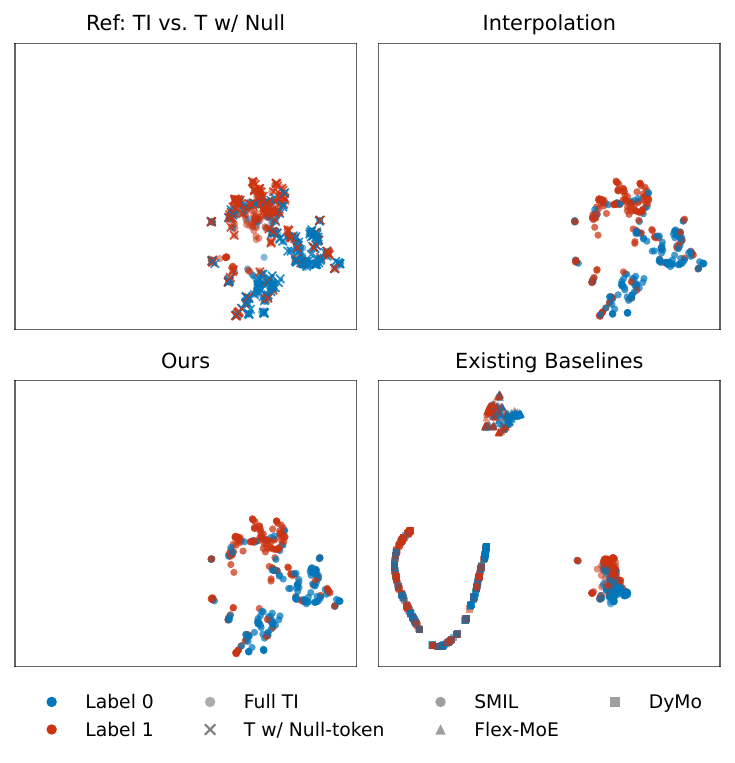}
    \caption{UMAP of fusion-layer representations.}
\end{subfigure}
\hfill
\begin{subfigure}{0.44\linewidth}
    \centering
    \includegraphics[width=\linewidth]{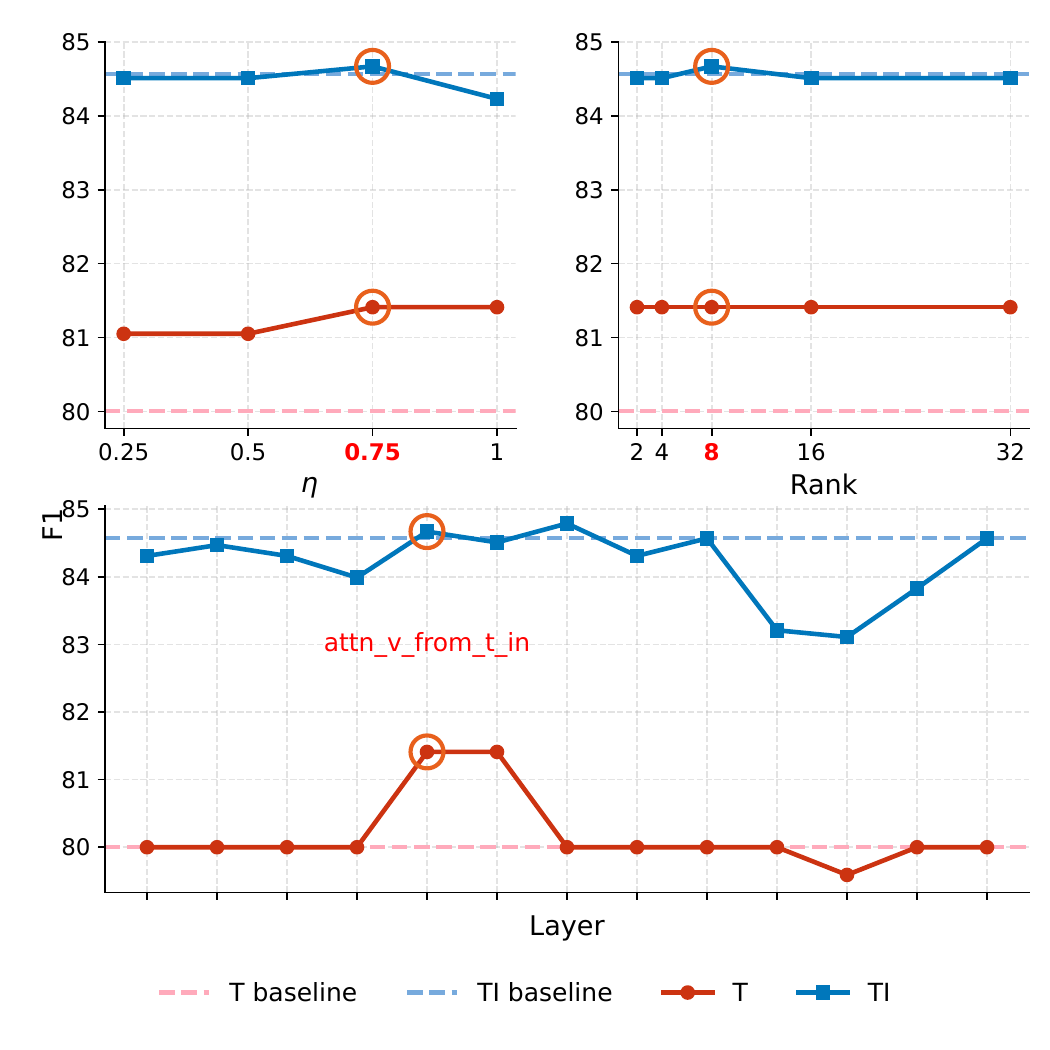}
    \caption{Ablations of $\eta$, rank, and target layer.}
\end{subfigure}
\caption{Representation structure and ablation analysis.
(a) On SNLI-VE, GU and layer interpolation preserve similar broad UMAP layouts.
(b) On IU-XRay, target-layer selection has the largest effect on robustness recovery.}
\label{fig:exp_repr_ablation}
\end{figure}

\noindent\textbf{Performance.}
Table~\ref{tab:exp_main_performance} reports mean F1 over three independent runs under the TI-T setting, where the image modality is entirely absent at inference. Corresponding sample standard deviations and supplementary ten-seed paired significance tests are provided in Appendix~\ref{app:performance_stability}. GU achieves the highest mean F1 among the compared methods on all three datasets, improving over the strongest baseline in each dataset by +2.38 on Hateful Memes, +1.27 on IU-XRay, and +2.21 on SNLI-VE. These consistent gains are obtained by editing only one selected layer after training. Thus, GU's advantage comes from directly addressing the failure mode identified in Section~\ref{sec:analysis}: harmful cross-modal dependencies in localized parameter subspaces.

The results also highlight why this deployment setting is challenging for data-side recovery and adaptive fusion methods. Many missing-modality methods are designed for settings with more available modalities and partial sample-level missingness, where the remaining modalities can still provide sufficient cross-modal information for recovery, routing, or adaptive fusion. In our setting, the system is bimodal and one of the two modalities is completely unavailable, leaving only text information. Under this stricter condition, baselines are less consistently effective: SMIL and Flex-MoE are competitive on SNLI-VE but do not consistently improve over the null-token baseline across datasets, while DyMo degrades substantially. By contrast, GU works from the model side, correcting the parameter subspace associated with harmful multimodal dependencies rather than attempting to reconstruct an unavailable modality. This explains why a simple null-token pathway can remain competitive on Hateful Memes and IU-XRay, while GU further improves it through targeted parameter-space correction.

\noindent\textbf{Representation-level evidence.}
Figure~\ref{fig:exp_repr_ablation}(a) visualizes fusion-layer representations on SNLI-VE using UMAP. GU and layer interpolation exhibit similar broad layouts, consistent with the limited scope of their single-layer edits. The quantitative analysis in Appendix~\ref{app:repr_evidence_details} shows that both edits retain high similarity to the original TI-T representations, while GU achieves larger error-margin gains and corrects more errors with a small increase in regressions. With its TI-T F1, these results support targeted task-relevant correction with substantial preservation of the representation structure.

Overall, the main results show consistent performance gains under complete missing-image inference, with GU ranking first among the methods in Table~\ref{tab:exp_main_performance} across all three datasets. Its advantage lies in correcting harmful learned dependencies within the model, rather than relying on data-side modality recovery or adaptive routing when insufficient cross-modal information remains. Comparisons with more baselines are in Appendix~\ref{app:additional_baselines}.

\subsection{Robustness--preservation trade-off across model families}
\label{sec:exp_preservation}

\begin{table}[t]
\caption{Family-level robustness--preservation trade-off on SNLI-VE. $\Delta T$ and $\Delta TI$ denote changes from the original TI model in TI-T and TI-TI, respectively.}
\label{tab:snlive_robustness}
\centering

\begin{tabular}{
l c
c >{\columncolor{cyan!10}}c
c >{\columncolor{cyan!10}}c
c >{\columncolor{cyan!10}}c
c >{\columncolor{cyan!10}}c
}
\toprule
Family & \#Var
& \multicolumn{2}{c}{TI-T}
& \multicolumn{2}{c}{TI-TI}
& \multicolumn{2}{c}{$\Delta$T}
& \multicolumn{2}{c}{$\Delta$TI} \\
\cmidrule(lr){3-4}
\cmidrule(lr){5-6}
\cmidrule(lr){7-8}
\cmidrule(lr){9-10}

&
& Interp. & \textbf{GU}
& Interp. & \textbf{GU}
& Interp. & \textbf{GU}
& Interp. & \textbf{GU} \\
\midrule

Fusion
& 3
& 71.34 & $\mathbf{73.86}$
& 86.33 & $\mathbf{86.37}$
& 0.98  & $\mathbf{3.49}$
& $-0.13$ & $\mathbf{-0.09}$ \\

Fusion (str.)
& 3
& 72.34 & $\mathbf{73.06}$
& 75.16 & $\mathbf{75.95}$
& 0.96  & $\mathbf{1.69}$
& $-1.13$ & $\mathbf{-0.34}$ \\

CLIP
& 2
& 74.50 & $\mathbf{74.68}$
& 76.13 & $\mathbf{76.70}$
& 1.04  & $\mathbf{1.22}$
& $-1.21$ & $\mathbf{-0.64}$ \\

VLM
& 1
& 70.65 & $\mathbf{70.65}$
& 81.72 & $\mathbf{81.73}$
& 0.01  & $\mathbf{0.01}$
& 0.00  & $\mathbf{+0.01}$ \\

\bottomrule
\end{tabular}
\end{table}

A key question is whether improving missing-modality robustness sacrifices full-modality capability. Table~\ref{tab:snlive_robustness} addresses this robustness--preservation trade-off by comparing TI-T and TI-TI performance across model families. GU improves TI-T performance across model families while keeping TI-TI performance nearly unchanged, indicating that the edited models do not collapse into text-only solutions. Compared with layer interpolation, GU usually yields larger TI-T gains and smaller TI-TI degradation, suggesting that subspace-level editing better preserves useful multimodal structure than direct parameter mixing. Complete per-variant results are provided in Appendix~\ref{app:preservation_tradeoff}.

The gain magnitude varies across architectures. GU is strongest for fusion models, where cross-modal dependencies are relatively localized in explicit fusion layers. Its gains are smaller for CLIP-style and BLIP architectures, where the relevant dependencies can be more distributed. Thus, GU is best viewed as a localized robustness correction method. It improves the robustness--preservation trade-off most when the harmful dependency is concentrated in the selected layer.

\subsection{Cost analysis}
\label{sec:exp_cost}

Beyond performance, we also assess GU's computational cost and deployment efficiency. We compare GU, post-hoc LoRA, and Dual CE on SNLI-VE using Fusion Cross-Attention, CLIP Mean Pooling, and BLIP-2 under the same hardware/software environment and evaluation protocol. LoRA starts from the same TI checkpoint as GU and adapts the same target layer with rank $8$. It trains on missing-image inputs for $4$ epochs at learning rate $10^{-4}$, selecting the best checkpoint on the development set. Dual CE retrains from scratch with joint full-input and missing-image supervision.

\begin{table*}[t]
\caption{Offline cost and robustness--preservation trade-off on SNLI-VE. Including reference training, GU uses less time and peak GPU memory than Dual CE.}
\label{tab:efficiency_cost}
\centering

\begin{tabular}{llrrrr}
\toprule
Architecture & Method / Phase
& Wall-clock (s)
& Peak Mem. (GiB)
& $\Delta T$
& $\Delta TI$ \\
\midrule

\multirow{5}{*}{Fusion CA}
& Post-hoc LoRA
& 1812.30
& 2.090
& +6.02
& -1.62 \\

& Dual CE
& 2592.03
& 13.954
& +13.57
& -0.94 \\

& T Reference
& 1849.09
& 5.241
& --
& -- \\

& \cellcolor{cyan!10}GU Edit
& \cellcolor{cyan!10}297.19
& \cellcolor{cyan!10}1.744
& \cellcolor{cyan!10}--
& \cellcolor{cyan!10}-- \\

& \cellcolor{cyan!10}\textbf{GU (Total)}
& \cellcolor{cyan!10}2146.28
& \cellcolor{cyan!10}5.241
& \cellcolor{cyan!10}+8.06
& \cellcolor{cyan!10}-0.11 \\

\midrule

\multirow{5}{*}{CLIP MP}
& Post-hoc LoRA
& 1910.39
& 4.423
& +1.03
& -3.66 \\

& Dual CE
& 2580.61
& 3.684
& +2.16
& -1.40 \\

& T Reference
& 1773.52
& 3.099
& --
& -- \\

& \cellcolor{cyan!10}GU Edit
& \cellcolor{cyan!10}230.17
& \cellcolor{cyan!10}3.104
& \cellcolor{cyan!10}--
& \cellcolor{cyan!10}-- \\

& \cellcolor{cyan!10}\textbf{GU (Total)}
& \cellcolor{cyan!10}2003.69
& \cellcolor{cyan!10}3.104
& \cellcolor{cyan!10}+1.43
& \cellcolor{cyan!10}-0.97 \\

\midrule

\multirow{5}{*}{BLIP-2}
& Post-hoc LoRA
& 2000.73
& 16.930
& +0.26
& -0.68 \\

& Dual CE
& 3079.53
& 32.857
& +4.41
& -0.53 \\

& T Reference
& 1993.90
& 24.957
& --
& -- \\

& \cellcolor{cyan!10}GU Edit
& \cellcolor{cyan!10}857.83
& \cellcolor{cyan!10}16.126
& \cellcolor{cyan!10}--
& \cellcolor{cyan!10}-- \\

& \cellcolor{cyan!10}\textbf{GU (Total)}
& \cellcolor{cyan!10}2851.73
& \cellcolor{cyan!10}24.957
& \cellcolor{cyan!10}+0.01
& \cellcolor{cyan!10}+0.01 \\

\bottomrule
\end{tabular}
\end{table*}

GU requires only a one-time edit before deployment, with no gradient updates to the TI checkpoint. As shown in Table~\ref{tab:efficiency_cost}, this stage takes $230$--$858$ seconds, reducing wall-clock time by $57$--$88\%$ relative to post-hoc LoRA. The resulting model introduces no additional parameters or forward-pass FLOPs relative to the original TI model. Inference uses only the edited checkpoint, without retaining the unimodal reference. For the complete offline comparison, costs start from an available TI checkpoint, excluding its original training cost. GU's total cost includes matched reference training and editing, whereas Dual CE includes its full retraining cost. The two GU phases run sequentially, so total runtime is their sum and peak memory is the larger of their two peaks. Even including reference training, GU uses $7$--$22\%$ less wall-clock time and $16$--$62\%$ less peak GPU memory than Dual CE across all three architectures.

Table~\ref{tab:efficiency_cost} also reports $\Delta T$ and $\Delta TI$ relative to the original TI model. GU combines lower total cost than Dual CE with better full-modality preservation. On Fusion and CLIP, it also achieves larger TI-T gains and smaller TI-TI losses than LoRA, although reference training makes its total offline runtime longer; on BLIP-2, the single-layer edit mainly preserves the original performance. Overall, GU provides an efficient post-training robustness correction with low offline cost, no additional inference overhead, and strong full-modality preservation.

\subsection{Ablation studies}
\label{sec:exp_ablation}

We ablate three key factors in GU: the geodesic coefficient $\eta$, the subspace rank $k$, and the target layer. As shown in Figure~\ref{fig:exp_repr_ablation}(b), GU is relatively stable across a range of $\eta$ and $k$ values, while the target layer plays a more decisive role. Varying $\eta$ from 0.25 to 1.0 causes only moderate changes in missing-modality performance and full-modality preservation, while ranks from 2 to 32 yield similar TI-T performance with small TI-TI fluctuations. We therefore select hyperparameters based on the development-set trade-off between TI-T improvement and TI-TI preservation.

Editing most layers yields limited gains, whereas applying GU to the selected cross-modal interaction layer \texttt{attn\_v\_from\_t\_in} produces the largest TI-T improvement while maintaining strong TI-TI performance. This aligns with Section~\ref{sec:analysis}, where subspace rotation and representational harm concentrate in cross-modal interaction stages. This result further supports the localized nature of missing-modality degradation, showing that a single well-chosen Grassmannian subspace edit can recover robustness without broadly modifying the model.

\section{Conclusion}

We systematically characterize a deployment-time failure mode in which full-modality multimodal models can underperform unimodal models when a modality is absent at inference. We show that this degradation is strongly associated with the principal subspace rotations of the cross-modal interaction layers and task-aware representational harm. Motivated by these findings, we propose Geodesic Unlearning, a lightweight post-training method that realigns a selected principal subspace, viewed as a point on the Grassmannian, toward a unimodal reference, with provably optimal alignment under a fixed subspace-distance budget. Experiments across model families demonstrate a favorable robustness--preservation trade-off, with further support from multi-choice VQA and missing-text evaluation. Overall, our theoretical and empirical results support a geometric view of the missing-modality problem and suggest localized subspace editing as a practical correction strategy.

\begin{ack}
This research was partially supported by the U.S. National Science Foundation under Award No. 2310260. Any opinions, findings, and conclusions or recommendations expressed in this paper are those of the authors and do not necessarily reflect the views of the National Science Foundation. The authors acknowledge the Center for Research Computing at Rice University for providing computational resources. The authors declare no competing interests.
\end{ack}

\bibliographystyle{plainnat}
\bibliography{references}


\clearpage
\appendix

\addtocontents{toc}{\protect\setcounter{tocdepth}{2}}

\renewcommand{\contentsname}{Appendix Contents}
\tableofcontents

\clearpage

\section{Limitations}
\label{app:limitations}

Despite the importance of our work in highlighting and addressing the structural causes of missing-modality degradation in multimodal learning, there are limitations worth noting. First, our evaluation is limited to classification tasks, including multiple-choice visual question answering. Extending GU to free-form multimodal generation is an important direction for future work. Second, concerns have been raised about rendered text in Hateful Memes and hypothesis-side artifacts in SNLI-VE. We address these concerns on the affected benchmarks through targeted preprocessing and complementary controls. Further evaluations on more datasets without these specific artifacts reproduce both the degradation and GU's improvements, confirming that neither finding requires their presence. Finally, although we cover fusion models, two-tower models, and VLMs across diverse domains, due to resource constraints, we do not conduct large-scale experiments on substantially larger models.

\section{Broader impacts}
\label{app:impacts}

This work improves the robustness of multimodal models to missing modality inputs, which is important in real-world settings where inputs may be incomplete due to sensor failure, privacy constraints, or deployment limitations. Such improvements may benefit applications like medical decision support and multimodal content understanding. However, increased robustness may encourage deployment in settings where incorrect predictions can still occur. In addition, robustness does not eliminate biases present in training data.

\section{Supplementary materials for empirical analysis}

\subsection{More results on missing-modality degradation}
\label{app:missing_modality_degradation}

The main text summarizes missing-modality degradation using aggregated bar plots. Table~\ref{tab:perf_gap_f1} provides complete F1 scores across model variants and datasets under full-modality evaluation (TI-TI), unimodal text evaluation (T-T), and deployment-time missing-image evaluation (TI-T). Results are mean $\pm$ sample standard deviation over three independent runs. IX and SV denote IU-XRay and SNLI-VE, respectively.

The reported mean scores support the trend in Section~\ref{sec:analysis_missing_failure}: $\mathrm{TI\text{-}TI} > \mathrm{T\text{-}T} > \mathrm{TI\text{-}T}$ holds in 26 of 27 settings, and $\mathrm{T\text{-}T} > \mathrm{TI\text{-}T}$ holds in all 27. The only non-strict case is strong-encoder Fusion Cross-Attention on HM, where $\mathrm{TI\text{-}TI} = \mathrm{T\text{-}T} = 66.93 > \mathrm{TI\text{-}T} = 66.85$. The effect is especially pronounced in models with explicit cross-modal interaction, such as cross-attention fusion on SNLI-VE, where the missing-modality gap is among the largest. These results support the view that deployment-time missing-modality degradation is associated not only with absent input information but also with cross-modal dependencies formed during multimodal training.

\subsection{More results on parameter subspace rotation}
\label{app:subspace_rotation}

To complement the main-text analysis of interaction-layer geometry, we provide a layer-wise comparison of fusion models on SNLI-VE. Table~\ref{tab:layerwise_subspace_snlive_fusion} reports input/output subspace overlap, mean principal angle, and Frobenius distance between each TI layer and its matched T reference.

Subspace shifts are nonuniform across layers. Projection and cross-modal interaction layers generally exhibit larger principal-angle deviations and lower subspace overlap than classifier layers. Cross-attention modules show large input- and output-side angle shifts, consistent with the main-text finding that interaction-layer geometry most strongly tracks the missing-modality gap. By contrast, the final classifier has near-perfect overlap and a near-zero output-side angle, suggesting that the dominant parameter-space changes arise before the final prediction layer.

These results support interaction-layer subspace rotation as a structural diagnostic for missing-modality degradation and a criterion for selecting localized GU editing targets.

\subsection{More results on representational harm under missing modalities}
\label{app:repr_harm}

The main text summarizes task-aware representational harm at the stage level. Table~\ref{tab:layerwise_repr_harm_snlive_fusion} provides layer-wise results for fusion models on SNLI-VE under missing-image evaluation. It reports the error--correct harm gap $\Delta_{\mathrm{harm}}$ and violation-rate gap $\Delta_{\mathrm{viol}}$. Larger gaps indicate that error samples occupy more task-inconsistent regions than correct samples.

Representational harm is nonuniform across layers. Harm signals become pronounced around multimodal interaction stages and often persist into the classifier input. This complements the parameter-subspace analysis in Table~\ref{tab:layerwise_subspace_snlive_fusion}. Image projection layers can exhibit large subspace rotations, but their harm scores are zero under missing-image evaluation because the image modality is unavailable. By contrast, cross-modal interaction layers show both substantial subspace rotation and nontrivial representational harm. These results motivate selecting target layers using both parameter subspace rotation and task-aware harm, rather than principal angles alone.

Together, these findings support the main-text interpretation that interaction-layer geometry is accompanied by task-relevant representation shifts under missing-modality inputs.

\begin{table*}[t]
\caption{F1 performance under three training--testing regimes. TI-T consistently underperforms T-T across model families and datasets.}
\label{tab:perf_gap_f1}
\centering
\small

\begin{tabular}{lllccc>{\columncolor{gray!10}}c}
\toprule
Family & Dataset & Variant & TI-TI & T-T & TI-T & $\Delta_{\mathrm{miss}}$ \\
\midrule

\multirow{9}{*}{Fusion}
& HM & Concat & $69.46\pm0.32$ & $68.99\pm0.25$ & $68.96\pm0.44$ & $0.50\pm0.14$ \\
& HM & Gate & $67.98\pm0.28$ & $67.86\pm0.23$ & $67.57\pm0.38$ & $0.41\pm0.12$ \\
& HM & Cross-Attn & $68.82\pm0.48$ & $68.41\pm0.40$ & $57.63\pm0.88$ & $11.19\pm0.45$ \\
& IX & Concat & $86.41\pm0.22$ & $81.94\pm0.18$ & $80.70\pm0.38$ & $5.71\pm0.18$ \\
& IX & Gate & $82.79\pm0.20$ & $81.18\pm0.17$ & $80.76\pm0.33$ & $2.03\pm0.15$ \\
& IX & Cross-Attn & $84.57\pm0.28$ & $80.97\pm0.23$ & $80.00\pm0.47$ & $4.57\pm0.22$ \\
& SV & Concat & $86.22\pm0.07$ & $81.84\pm0.05$ & $73.04\pm0.24$ & $13.18\pm0.18$ \\
& SV & Gate & $86.53\pm0.08$ & $82.31\pm0.06$ & $71.70\pm0.29$ & $14.83\pm0.22$ \\
& SV & Cross-Attn & $86.64\pm0.10$ & $81.85\pm0.08$ & $66.35\pm0.38$ & $20.29\pm0.30$ \\

\midrule

\multirow{9}{*}{\shortstack[l]{Fusion\\(strong)}}
& HM & Concat & $67.49\pm0.30$ & $67.12\pm0.24$ & $66.50\pm0.46$ & $0.99\pm0.18$ \\
& HM & Gate & $67.12\pm0.34$ & $66.85\pm0.27$ & $65.51\pm0.53$ & $1.61\pm0.22$ \\
& HM & Cross-Attn & $66.93\pm0.22$ & $66.93\pm0.18$ & $66.85\pm0.24$ & $0.08\pm0.03$ \\
& IX & Concat & $81.51\pm0.21$ & $80.35\pm0.16$ & $78.06\pm0.36$ & $3.45\pm0.17$ \\
& IX & Gate & $79.93\pm0.18$ & $79.79\pm0.14$ & $79.24\pm0.25$ & $0.69\pm0.08$ \\
& IX & Cross-Attn & $80.14\pm0.24$ & $79.46\pm0.19$ & $77.78\pm0.38$ & $2.36\pm0.16$ \\
& SV & Concat & $76.44\pm0.06$ & $75.99\pm0.05$ & $72.78\pm0.19$ & $3.66\pm0.14$ \\
& SV & Gate & $72.72\pm0.07$ & $71.51\pm0.06$ & $67.65\pm0.24$ & $5.07\pm0.18$ \\
& SV & Cross-Attn & $79.71\pm0.08$ & $79.23\pm0.07$ & $73.69\pm0.27$ & $6.02\pm0.20$ \\

\midrule

\multirow{6}{*}{CLIP}
& HM & Late Concat & $67.31\pm0.27$ & $67.21\pm0.22$ & $66.58\pm0.41$ & $0.73\pm0.16$ \\
& HM & Mean Pooling & $67.14\pm0.20$ & $66.95\pm0.17$ & $66.87\pm0.24$ & $0.27\pm0.05$ \\
& IX & Late Concat & $80.85\pm0.19$ & $80.40\pm0.15$ & $79.45\pm0.29$ & $1.40\pm0.12$ \\
& IX & Mean Pooling & $78.00\pm0.20$ & $77.97\pm0.16$ & $74.30\pm0.38$ & $3.70\pm0.20$ \\
& SV & Late Concat & $78.05\pm0.06$ & $78.01\pm0.05$ & $72.77\pm0.21$ & $5.28\pm0.16$ \\
& SV & Mean Pooling & $76.62\pm0.06$ & $76.52\pm0.05$ & $74.15\pm0.18$ & $2.47\pm0.13$ \\

\midrule

\multirow{3}{*}{\shortstack[l]{VLM\\(BLIP-2)}}
& HM & BLIP-2 & $68.85\pm0.38$ & $66.58\pm0.31$ & $66.21\pm0.59$ & $2.64\pm0.24$ \\
& IX & BLIP-2 & $81.74\pm0.26$ & $79.38\pm0.22$ & $78.00\pm0.46$ & $3.74\pm0.23$ \\
& SV & BLIP-2 & $81.72\pm0.18$ & $77.51\pm0.07$ & $70.64\pm0.70$ & $11.08\pm0.66$ \\

\bottomrule
\end{tabular}
\end{table*}

\clearpage

\begin{table*}[p]
\caption{Layer-wise task-aware representational harm for fusion models on SNLI-VE. Harm gaps appear at cross-modal interaction stages and often persist near the classifier.}
\label{tab:layerwise_repr_harm_snlive_fusion}
\centering
\small
\begin{tabular}{llcc}
\toprule
Variant & Layer & $\Delta_{\mathrm{harm}}$ & $\Delta_{\mathrm{viol}}$ \\
\midrule

\multirow{5}{*}{Concat}
& text\_proj\_out & 0.5187 & 0.6170 \\
& image\_proj\_out & 0.0000 & 0.0000 \\
& fusion\_mlp0\_concat\_input & 0.5187 & 0.6170 \\
& fusion\_mlp3\_output & 0.5617 & 0.7308 \\
& head\_last\_linear\_input & 0.5617 & 0.7308 \\
\midrule

\multirow{8}{*}{Gate}
& text\_proj\_out & 0.3881 & 0.5735 \\
& image\_proj\_out & 0.0000 & 0.0000 \\
& fusion\_mlp0\_concat\_input & 0.3881 & 0.5735 \\
& fusion\_mlp3\_h\_base & 0.2373 & 0.5992 \\
& gate\_mlp0\_gate & 0.1066 & 0.5617 \\
& fusion\_pre\_ln & 0.2996 & 0.6127 \\
& fusion\_output & 0.2996 & 0.6113 \\
& head\_last\_linear\_input & 0.2996 & 0.6113 \\
\midrule

\multirow{13}{*}{Cross-Attn}
& text\_proj\_out & 0.4585 & 0.4339 \\
& image\_proj\_out & 0.0000 & 0.0000 \\
& attn\_t\_from\_v\_in\_proj\_to\_text\_pooled & 0.4134 & 0.4464 \\
& attn\_t\_from\_v\_out\_proj\_to\_text\_pooled & 0.4134 & 0.4464 \\
& attn\_v\_from\_t\_in\_proj\_to\_image\_pooled & 0.3673 & 0.4557 \\
& attn\_v\_from\_t\_out\_proj\_to\_image\_pooled & 0.3673 & 0.4557 \\
& ffn\_t\_0\_to\_text\_pooled & 0.4134 & 0.4464 \\
& ffn\_t\_3\_to\_text\_pooled & 0.4134 & 0.4464 \\
& ffn\_v\_0\_to\_image\_pooled & 0.3673 & 0.4557 \\
& ffn\_v\_3\_to\_image\_pooled & 0.3673 & 0.4557 \\
& combine\_mlp0\_concat\_input & 0.3896 & 0.4420 \\
& combine\_mlp3\_output & 0.4445 & 0.5868 \\
& head\_last\_linear\_input & 0.4445 & 0.5868 \\

\bottomrule
\end{tabular}
\end{table*}

\begin{table*}[t]
\caption{Layer-wise parameter subspace rotation for fusion models on SNLI-VE. Projection and cross-modal interaction layers generally exhibit larger rotations than the final classifier.}
\label{tab:layerwise_subspace_snlive_fusion}
\centering
\small
\begin{tabular}{llccc}
\toprule
Variant & Layer & Overlap (In / Out) & Mean Angle (In / Out) & Fro. Dist. \\
\midrule

\multirow{5}{*}{Concat}
& text\_proj & 0.9666 / 0.9666 & 8.3333 / 8.3150 & 1.4884 \\
& image\_proj & 0.8368 / 0.8409 & 19.2640 / 18.7282 & 2.5745 \\
& fusion\_mlp0 & 0.9307 / 0.9314 & 11.4472 / 11.3110 & 1.8050 \\
& fusion\_mlp3 & 0.9578 / 0.9573 & 4.2398 / 4.2895 & 1.3692 \\
& head\_last\_linear & 0.9991 / 1.0000 & 1.7072 / 0.0000 & 0.0511 \\
\midrule

\multirow{6}{*}{Gate}
& text\_proj & 0.9715 / 0.9714 & 7.7292 / 7.7325 & 1.4133 \\
& image\_proj & 0.8615 / 0.8656 & 17.8397 / 17.2876 & 2.4042 \\
& fusion\_mlp0 & 0.9129 / 0.9126 & 12.3658 / 12.2981 & 1.8239 \\
& fusion\_mlp3 & 0.9249 / 0.9254 & 8.3895 / 8.2923 & 1.4560 \\
& gate\_mlp0 & 0.9408 / 0.9425 & 10.8734 / 10.5926 & 2.0828 \\
& head\_last\_linear & 0.9994 / 1.0000 & 1.3227 / 0.0000 & 0.0367 \\
\midrule

\multirow{13}{*}{Cross-Attn}
& text\_proj & 0.9656 / 0.9658 & 8.1285 / 8.0749 & 1.4191 \\
& image\_proj & 0.8777 / 0.8818 & 16.1052 / 15.5300 & 2.3401 \\
& attn\_t\_from\_v\_in\_proj & 0.9152 / 0.9127 & 13.3996 / 13.8179 & 2.3295 \\
& attn\_t\_from\_v\_out\_proj & 0.8701 / 0.8666 & 17.2543 / 17.6767 & 1.4386 \\
& attn\_v\_from\_t\_in\_proj & 0.8845 / 0.8796 & 16.0031 / 16.5666 & 2.4871 \\
& attn\_v\_from\_t\_out\_proj & 0.9315 / 0.9312 & 10.2142 / 10.2819 & 0.9027 \\
& ffn\_t\_0 & 0.9873 / 0.9870 & 5.2286 / 5.3021 & 1.1664 \\
& ffn\_t\_3 & 0.9862 / 0.9860 & 5.0457 / 5.0703 & 1.0904 \\
& ffn\_v\_0 & 0.9771 / 0.9767 & 6.4555 / 6.4909 & 1.2853 \\
& ffn\_v\_3 & 0.9881 / 0.9882 & 4.6682 / 4.6642 & 1.0396 \\
& combine\_mlp0 & 0.9638 / 0.9642 & 8.4807 / 8.4070 & 1.5600 \\
& combine\_mlp3 & 0.9471 / 0.9464 & 5.3530 / 5.4254 & 1.0109 \\
& head\_last\_linear & 0.9988 / 1.0000 & 1.9483 / 0.0000 & 0.0501 \\

\bottomrule
\end{tabular}
\end{table*}

\clearpage

\section{Methodology details}
\label{app:gu_details}

\subsection{Grassmannian geodesic construction}
\label{app:gu_geodesic_details}

Intuitively, the construction pairs the principal directions of the multimodal and unimodal subspaces and rotates each multimodal direction toward its paired reference direction. A single edit strength specifies the same fraction of angular correction for every pair. The resulting basis is then expressed in the original TI coordinates for reconstruction with the retained output factor, principal coefficients, and residual. This section provides the explicit Grassmannian geodesic used in Eq.~\eqref{eq:gu_geodesic_main}. 
Let $V_{\mathrm{TI},k},V_{\mathrm{T},k}\in\mathbb{R}^{d_i\times k}$ be orthonormal bases of the top-$k$ input subspaces of the multimodal model and the matched unimodal reference. 
We first compute the compact SVD of their subspace overlap:
\begin{equation}
V_{\mathrm{TI},k}^{\top}V_{\mathrm{T},k}
=
P\cos(\Theta)Q^\top ,
\label{eq:app_gu_overlap_svd}
\end{equation}
where the diagonal entries $\theta_i\in[0,\pi/2]$ of $\Theta$ are the principal angles between the two subspaces. 
We then form the aligned principal bases
\begin{equation}
A = V_{\mathrm{TI},k}P,
\qquad
B = V_{\mathrm{T},k}Q.
\label{eq:app_gu_principal_bases}
\end{equation}
The geodesic direction is computed as
\begin{equation}
Y
=
(B-A\cos\Theta)(\sin\Theta)^{\dagger},
\label{eq:app_gu_direction}
\end{equation}
where $(\cdot)^{\dagger}$ denotes the Moore--Penrose pseudoinverse. For $\theta_i=0$, the common direction is unchanged and the corresponding column of $Y$ is zero. 
The geodesic basis in principal coordinates and its expression in the original TI singular-vector coordinates are
\begin{equation}
\widetilde V(\eta)
=
A\cos(\eta\Theta)
+
Y\sin(\eta\Theta),
\qquad
V(\eta)=\widetilde V(\eta)P^\top,
\quad \eta\in[0,1].
\label{eq:app_gu_geodesic}
\end{equation}
Here $\widetilde V(0)=A$ and $\widetilde V(1)=B$. Multiplication by $P^\top$ preserves the subspace and orthonormality while expressing the basis in the TI coordinates used for reconstruction. Consequently, $V(0)=V_{\mathrm{TI},k}$ and $V(1)=V_{\mathrm T,k}QP^\top$, so the path starts at the original TI basis and ends in the reference subspace.

\subsection{Geometric rationale and path comparison}
\label{app:gu_geometric_rationale}

\noindent\textbf{The edited object.}
The top-$k$ right singular subspace specifies the dominant input directions read by the layer. Sections~\ref{sec:analysis_para_subspace} and~\ref{sec:analysis_repr_harm} motivate targeting this orientation through parameter deviation and task-aware representation harm. Its Grassmannian representation identifies all orthonormal bases spanning the same subspace. GU retains the original output factor $U_{\mathrm{TI},k}$, principal coefficients $\Sigma_{\mathrm{TI},k}$, and residual $R_{\mathrm{TI}}$; direct weight interpolation changes these components together.

\noindent\textbf{Edit-budget optimality.} The geometric question is how close an edited subspace can get to the reference while limiting its departure from the original multimodal subspace. The edit budget specifies this allowed departure. A shortest path uses the budget fully toward alignment: every unit of movement reduces the remaining distance to the reference by one unit.

Let $S_{\mathrm{TI}}=\operatorname{span}(V_{\mathrm{TI},k})$ and $S_{\mathrm T}=\operatorname{span}(V_{\mathrm T,k})$. We use the canonical Grassmann geodesic distance~\cite{edelman1998geometry},
\begin{equation}
d(S_1,S_2)=\left(\sum_{i=1}^{k}\theta_i(S_1,S_2)^2\right)^{1/2},
\qquad D=d(S_{\mathrm{TI}},S_{\mathrm T}).
\label{eq:app_gu_distance}
\end{equation}
This distance defines the edit budget; the mean principal angle in Section~\ref{sec:analysis_para_subspace} is the diagnostic summary. The subspace path in Eq.~\eqref{eq:app_gu_geodesic} can be written intrinsically as
\begin{equation}
S_\eta=\operatorname{Exp}_{S_{\mathrm{TI}}}\!\left(
\eta\,\operatorname{Log}_{S_{\mathrm{TI}}}(S_{\mathrm T})\right),
\qquad 0\leq\eta\leq1,
\label{eq:app_gu_exp_log}
\end{equation}
where $\operatorname{Log}$ denotes a minimizing logarithm; when it is nonunique, a minimizing branch is fixed. Each principal direction advances through $\eta\theta_i$, leaving $(1-\eta)\theta_i$ to the reference. Hence $d(S_{\mathrm{TI}},S_\eta)=\eta D$ and $d(S_\eta,S_{\mathrm T})=(1-\eta)D$.
For any candidate $S$ satisfying $d(S_{\mathrm{TI}},S)\leq\eta D$, the triangle inequality gives
\begin{equation}
d(S,S_{\mathrm T})\geq D-d(S_{\mathrm{TI}},S)\geq(1-\eta)D.
\label{eq:app_gu_budget_bound}
\end{equation}
The geodesic point $S_\eta$ attains this bound, proving Eq.~\eqref{eq:gu_edit_budget}. Thus, for a fixed subspace-distance budget, GU makes the greatest possible progress toward the matched reference. This optimality concerns the geometric path; it does not assert optimal task performance.

\noindent\textbf{Comparison with aligned linear interpolation.}
Using the same principal bases $A,B$, consider the extrinsic path
\begin{equation}
L(\eta)=\operatorname{polar}\!\left((1-\eta)A+\eta B\right),
\label{eq:app_gu_linear_polar}
\end{equation}
where $\operatorname{polar}(X)=X(X^\top X)^{-1/2}$ is the orthonormal polar factor. Write $a_i,b_i,y_i$ for corresponding columns of $A,B,Y$. Since $b_i=\cos\theta_i\,a_i+\sin\theta_i\,y_i$, the $i$th interpolated column is
\begin{equation}
\ell_i(\eta)=
\frac{(1-\eta+\eta\cos\theta_i)a_i+\eta\sin\theta_i\,y_i}
{\sqrt{(1-\eta+\eta\cos\theta_i)^2+\eta^2\sin^2\theta_i}}.
\label{eq:app_gu_linear_direction}
\end{equation}
The unnormalized columns are mutually orthogonal, so the polar step normalizes each separately. Its displacement from $a_i$ within the corresponding principal plane is therefore
\begin{equation}
\phi_i(\eta)=\operatorname{atan2}\!\left(
\eta\sin\theta_i,\,1-\eta+\eta\cos\theta_i\right).
\label{eq:app_gu_linear_angle}
\end{equation}
In general, $\phi_i(\eta) \neq \eta\theta_i$. The distinction is the meaning of the interpolation coefficient: mixing all aligned basis vectors with the same coefficient does not generally advance them through the same fraction of their respective angles. GU gives $\eta$ a consistent geometric meaning across directions, advancing every principal angle proportionally at constant geodesic speed. For weight reconstruction, the corresponding linear-path basis is $L(\eta)P^\top$, using the same TI coordinates as GU.

\subsection{Implementation notes}
\label{app:gu_implementation_notes}

In implementation, the singular values of $V_{\mathrm{TI},k}^{\top}V_{\mathrm{T},k}$ are clamped to $[-1,1]$ before applying $\arccos$ to compute the principal angles. 
For angles close to zero, the corresponding entries of $(\sin\Theta)^{\dagger}$ are stabilized with a small numerical constant. 
After computing $V(\eta)$, the edited layer is reconstructed using Eq.~\eqref{eq:gu_reconstruction} in the main text:
\begin{equation}
W'
=
R_{\mathrm{TI}}
+
U_{\mathrm{TI},k}\Sigma_{\mathrm{TI},k}V(\eta)^\top .
\label{eq:app_gu_reconstruction}
\end{equation}
The conversion in Eq.~\eqref{eq:app_gu_geodesic} ensures that $W'=W_{\mathrm{TI}}$ when $\eta=0$. All other layers remain unchanged.

\section{Supplementary materials for experimental setup}
\label{app:experimental_setup}

\subsection{Dataset}
\label{app:dataset_details}

\begin{table}[h]
\caption{Dataset split statistics used in our experiments.}
\label{tab:dataset_statistics}
\centering
\begin{tabular}{lccc}
\toprule
Split & Hateful Memes & SNLI-VE & IU-XRay \\
\midrule
Train & 8,500 & 353,482 & 2,928 \\
Dev   & 500   & 11,898  & 366 \\
Test  & 1,000 & 11,937  & 366 \\
\bottomrule
\end{tabular}
\end{table}

Our main experiments use three text-image multimodal benchmarks: \textit{Hateful Memes}~\cite{kiela2020hateful}, \textit{SNLI-VE}~\cite{xie2019visual}, and \textit{IU-XRay}~\cite{demner2016preparing}. \textit{Hateful Memes} is a social multimodal classification benchmark that requires jointly reasoning over meme images and associated text. \textit{SNLI-VE} is a visual entailment benchmark that pairs images with textual hypotheses. \textit{IU-XRay} is a medical vision-language dataset constructed from chest X-ray images and radiology reports. For consistency, they are formulated as binary classification tasks. For \textit{SNLI-VE}, which originally comprises entailment, contradiction, and neutral classes, we remove neutral examples. This maintains a consistent binary classification protocol across datasets while mitigating the impact of documented labeling issues associated with the neutral class~\cite{gururangan-etal-2018-annotation, do2021esnlivecorrectedvisualtextualentailment}. Table~\ref{tab:dataset_statistics} summarizes the dataset splits used in our experiments. We use the development set for hyperparameter selection and report final results on the held-out test set.

\subsection{Baselines}
\label{app:baselines}

For experimental organization, we group the compared methods into multimodal-specific methods, general training and adaptation baselines, and parameter-editing controls. This grouping reflects each method's role and applicability in the evaluation, rather than a mutually exclusive taxonomy of prior work. Coverage follows each method's architectural requirements: SMIL, Flex-MoE, and DyMo are compared under a matched fusion backbone, while additional architecture-compatible methods extend the evaluation to other model families.

\subsubsection{Multimodal-specific methods}
\label{app:missing_modality_baseline_details}

\noindent\textbf{SMIL, Flex-MoE, and DyMo.}
We evaluate SMIL~\cite{ma2021smil}, Flex-MoE~\cite{yun2024flex}, and DyMo~\cite{du2026inferencetime} on all three main datasets using the unified RoBERTa--ResNet50 cross-attention fusion backbone.
This choice follows the structural assumptions of their released implementations.
SMIL couples Bayesian meta-learning with missing-modality feature reconstruction, requiring choices of reconstruction targets and auxiliary-network insertion points.
Flex-MoE uses a missing-modality bank, experts, and specialized routers; DyMo uses recovered-modality candidates and inference-time modality selection.
Extending these methods to CLIP-style or BLIP-2 models is possible, but would require method-specific changes to reconstruction targets, native fusion or Q-Former components, or routing and recovery modules.
Such redesigns would alter the architecture, parameter budget, and optimization pipeline, adding design choices to matched-backbone comparisons.
This setting provides a common backbone and training protocol while retaining each method's auxiliary components.

For SMIL, Flex-MoE, and DyMo, we use the official codebases released by the original papers and keep unspecified settings consistent with their default configurations. To ensure a fair comparison, we use the same number of training epochs across our models: 10 for Hateful Memes, 10 for IU-XRay, and 4 for SNLI-VE. We also match the main learning rates of these baselines to our unified cross-attention fusion setting. The only exception is the SMIL inner-loop learning rate, which we set to $1.0\times10^{-3}$ to follow the official implementation.

\noindent\textbf{MoRA and OGM-GE.}
MoRA~\cite{zhao2026mora} provides a missing-modality adaptation baseline on CLIP Late Concat and Mean Pooling across the three main datasets.
Its modality-specific and shared low-rank updates couple paired text and vision encoder weights; applying this formulation to Fusion Concat would require new insertion points and a fusion-specific sharing rule. OGM-GE~\cite{peng2022balanced} provides a training-time modality-balancing baseline on Fusion Concat and both CLIP variants across the same datasets.
CLIP Mean Pooling permits an exact decomposition of the full logit into text and image contributions.
For nonlinear Late Concat, modality confidence is estimated through native unimodal forward paths while retaining the published discrepancy, gradient-modulation, and generalization-enhancement rules.
Results are reported in Appendix~\ref{app:mora_ogm}.

SiMO~\cite{wen2026simo} uses a specialized collaborative-perception architecture and staged training. Porting it to our setting would replace the evaluated backbone and training pipeline, rather than provide a matched-backbone comparison.

\subsubsection{General training and adaptation baselines}

\noindent\textbf{Dual CE.}
On SNLI-VE, we retrain Fusion Cross-Attention, CLIP Mean Pooling, and BLIP-2 from scratch with a shared classifier jointly supervising full-input and missing-image paths:
\[
\mathcal{L}_{\mathrm{Dual\,CE}} = \mathcal{L}_{\mathrm{TI}} + 0.5\,\mathcal{L}_{\mathrm{T}}.
\]
Both terms use binary cross-entropy with logits, evaluated with full text--image inputs and with the image replaced by a null input, respectively.
Dual CE+GU applies GU to the resulting checkpoint to test whether the localized edit complements joint supervision.
The comparison is reported in Appendix~\ref{app:dual_ce}.

\noindent\textbf{LoRA.}
For the SNLI-VE comparison on the same three architectures, post-hoc LoRA~\cite{hu2021loralowrankadaptationlarge} starts from the same TI checkpoint as GU and adapts the same target layer on missing-image inputs.
We use rank $8$, train for $4$ epochs at learning rate $10^{-4}$, and select the best checkpoint on the development set.
Performance and computational costs are reported in Section~\ref{sec:exp_cost}.

\subsubsection{Parameter-editing controls}
\label{app:layer_interpolation_details}

\noindent\textbf{Weight (layer interpolation).}
For layer interpolation, we use the same target layer as Geodesic Unlearning for each architecture variant, as listed in Table~\ref{tab:app_gu_target_layers}. Given the selected multimodal parameter matrix $W_{\mathrm{TI}}$ and the corresponding unimodal reference parameter matrix $W_{\mathrm{T}}$, layer interpolation constructs the edited parameter as
\[
W' = (1-\alpha)W_{\mathrm{TI}} + \alpha W_{\mathrm{T}} .
\]
We fix $\alpha=0.5$ for all datasets and models. This baseline directly mixes parameter values, whereas GU edits only the selected Grassmannian subspace while preserving the residual parameter structure.

\noindent\textbf{Linear (subspace interpolation).}
In the Fusion Cross-Attention controls on the three main datasets, Linear interpolates the aligned principal bases and applies polar retraction, as defined in Eq.~\eqref{eq:app_gu_linear_polar}.
It uses the same target layer, matched checkpoints and reference, rank $k=8$, and coefficient $\eta=0.75$ as GU, while retaining the original TI output factor, principal coefficients, and residual through the reconstruction coordinates in Appendix~\ref{app:gu_geometric_rationale}.
These controls isolate the effect of the edited object and the subspace path; their results are reported in Appendix~\ref{app:controlled_interventions}.

\subsection{Compute resources}
\label{app:compute_resources}

All experiments can be run on a single NVIDIA A100 GPU with 80GB of memory. This includes training the multimodal and unimodal reference models, running Geodesic Unlearning and layer interpolation, and evaluating the missing-modality baselines. Geodesic Unlearning itself is applied as a post-training single-layer parameter edit and therefore adds negligible computational overhead compared with model training.

\subsection{Implementation}
\label{app:implementation_details}

\subsubsection{Backbone configurations}
\label{app:backbone_configs}

For model families in Table~\ref{tab:analysis_models}, we use the following backbone configurations. Basic fusion models use RoBERTa and ResNet50, while the stronger fusion setting uses Qwen2.5-3B and OpenCLIP ViT-bigG-14. CLIP-style two-tower models use OpenCLIP ViT-L-14, and VLM experiments use BLIP-2 with a classifier head. For each TI model, we use a unimodal reference with the same architecture and pretrained parameterization. GU requires dimensionally and semantically corresponding layers; references without this correspondence would require an additional alignment map.

\subsubsection{Hyperparameters}
\label{app:gu_hyperparams}

For the three binary benchmarks, standard TI training uses a shared classifier and binary cross-entropy with logits on full text--image inputs. Dual CE adds missing-image supervision as specified in Appendix~\ref{app:baselines}.

For all models used with Geodesic Unlearning, we train the underlying multimodal models with weight decay $1.0\times 10^{-2}$ and select the best checkpoint on the development set. The maximum number of training epochs is 10 for Hateful Memes, 10 for IU-XRay, and 4 for SNLI-VE. Tables~\ref{tab:app_lr_basic_fusion}--\ref{tab:app_lr_blip} summarize the TI-mode learning rates used for each model family. The symbol ``--'' indicates that the module is not present in the corresponding configuration. For basic fusion models, no module-specific learning rates are used, and a single global learning rate of $3.0\times 10^{-5}$ is applied to all trainable parameters, including unfrozen encoder backbones.

\subsubsection{Intervention details}
\label{app:gu_intervention_details}

For Geodesic Unlearning, we select the geodesic coefficient $\eta=0.75$ and subspace rank $k=8$ on the development set and keep both fixed across all datasets and model families.
For each architecture variant, candidate layers are chosen from the interaction and nearby projection/classifier layers analyzed in Section~\ref{sec:analysis}.
The target layer is selected on the development set using the angle--harm score in Eq.~\eqref{eq:target_layer_score}, which combines the principal-angle deviation between the TI model and the T reference with the task-aware harm gap under missing-modality evaluation.
When a task-aware harm score is measured at the functional-stage level rather than at an individual layer, we assign the same stage-level harm score to all candidate layers within that stage.
The test set is never used for target-layer selection.
Table~\ref{tab:app_gu_target_layers} lists the selected target layer for each architecture variant.

\begin{table}[h]
\caption{Learning rates for missing-modality baselines. Unspecified hyperparameters follow the official codebase defaults.}
\label{tab:app_baseline_lrs}
\centering
\begin{tabular}{lccccc}
\toprule
Dataset & \textbf{GU} & SMIL Outer & SMIL Inner & Flex-MoE & DyMo \\
\midrule
Hateful Memes & $3.0{\times}10^{-5}$ & $3.0{\times}10^{-5}$ & $1.0{\times}10^{-3}$ & $3.0{\times}10^{-5}$ & $3.0{\times}10^{-5}$ \\
SNLI-VE       & $3.0{\times}10^{-5}$ & $3.0{\times}10^{-5}$ & $1.0{\times}10^{-3}$ & $3.0{\times}10^{-5}$ & $3.0{\times}10^{-5}$ \\
IU-XRay      & $3.0{\times}10^{-5}$ & $3.0{\times}10^{-5}$ & $1.0{\times}10^{-3}$ & $3.0{\times}10^{-5}$ & $3.0{\times}10^{-5}$ \\
\bottomrule
\end{tabular}
\end{table}

\begin{table}[h]
\caption{Learning rates for basic fusion models. A single global learning rate is used for all trainable parameters.}
\label{tab:app_lr_basic_fusion}
\centering
\begin{tabular}{lccc}
\toprule
Dataset & Concat & Cross-Attention & Gate \\
\midrule
Hateful Memes & $3.0{\times}10^{-5}$ & $3.0{\times}10^{-5}$ & $3.0{\times}10^{-5}$ \\
SNLI-VE       & $3.0{\times}10^{-5}$ & $3.0{\times}10^{-5}$ & $3.0{\times}10^{-5}$ \\
IU-XRay      & $3.0{\times}10^{-5}$ & $3.0{\times}10^{-5}$ & $3.0{\times}10^{-5}$ \\
\bottomrule
\end{tabular}
\end{table}

\begin{table}[h]
\caption{Module-wise learning rates for fusion models with stronger encoders. Text/image denotes both \texttt{text\_encoder.proj} and \texttt{image\_encoder.proj}; bottleneck/head denotes both modules.}
\label{tab:app_lr_strong_fusion}
\centering
\begin{tabular}{llccc}
\toprule
Dataset & Variant & Text/Image & Fusion & Bottleneck/Head \\
\midrule
\multirow{3}{*}{Hateful Memes}
& Concat          & $2.0{\times}10^{-5}$ & $3.0{\times}10^{-5}$ & $3.0{\times}10^{-5}$ \\
& Cross-Attention & $2.0{\times}10^{-5}$ & $2.0{\times}10^{-5}$ & $3.0{\times}10^{-5}$ \\
& Gate           & $1.0{\times}10^{-5}$ & $2.0{\times}10^{-5}$ & $2.0{\times}10^{-5}$ \\
\midrule
\multirow{3}{*}{SNLI-VE}
& Concat          & $2.0{\times}10^{-5}$ & $3.0{\times}10^{-5}$ & $3.0{\times}10^{-5}$ \\
& Cross-Attention & $1.0{\times}10^{-5}$ & $1.0{\times}10^{-5}$ & $1.0{\times}10^{-5}$ \\
& Gate           & $2.0{\times}10^{-5}$ & $3.0{\times}10^{-5}$ & $3.0{\times}10^{-5}$ \\
\midrule
\multirow{3}{*}{IU-XRay}
& Concat          & $2.0{\times}10^{-5}$ & $3.0{\times}10^{-5}$ & $3.0{\times}10^{-5}$ \\
& Cross-Attention & $2.0{\times}10^{-5}$ & $2.0{\times}10^{-5}$ & $3.0{\times}10^{-5}$ \\
& Gate           & $1.0{\times}10^{-5}$ & $2.0{\times}10^{-5}$ & $2.0{\times}10^{-5}$ \\
\bottomrule
\end{tabular}
\end{table}

\begin{table}[h]
\caption{Module-wise learning rates for CLIP-style two-tower models. Text/image denotes both projection modules; bottleneck/head denotes both modules.}
\label{tab:app_lr_clip}
\centering
\begin{tabular}{llccc}
\toprule
Dataset & Variant & Text/Image & Fusion & Bottleneck/Head \\
\midrule
\multirow{2}{*}{Hateful Memes}
& Late Concat  & $1.0{\times}10^{-5}$ & $1.0{\times}10^{-5}$ & $1.0{\times}10^{-5}$ \\
& Mean Pooling & $1.0{\times}10^{-5}$ & -- & $1.0{\times}10^{-5}$ \\
\midrule
\multirow{2}{*}{SNLI-VE}
& Late Concat  & $1.0{\times}10^{-5}$ & $1.0{\times}10^{-5}$ & $1.0{\times}10^{-5}$ \\
& Mean Pooling & $1.0{\times}10^{-5}$ & -- & $1.0{\times}10^{-5}$ \\
\midrule
\multirow{2}{*}{IU-XRay}
& Late Concat        & $1.0{\times}10^{-5}$ & $1.0{\times}10^{-5}$ & $1.0{\times}10^{-5}$ \\
& Mean Pooling       & $1.0{\times}10^{-5}$ & -- & $1.0{\times}10^{-5}$ \\
\bottomrule
\end{tabular}
\end{table}

\begin{table}[h]
\caption{Module-wise learning rates for BLIP-2 based VLM models. Q-Former denotes both \texttt{vlm\_encoder.qformer\_embeddings} and \texttt{vlm\_encoder.backbone.qformer}; projection denotes both \texttt{vlm\_encoder.backbone.language\_projection} and \texttt{vlm\_encoder.proj}.}
\label{tab:app_lr_blip}
\centering
\begin{tabular}{lcccc}
\toprule
Dataset & Q-Former & Projection & Bottleneck & Head \\
\midrule
Hateful Memes & $5.0{\times}10^{-6}$ & $5.0{\times}10^{-6}$ & $2.0{\times}10^{-5}$ & $2.0{\times}10^{-5}$ \\
SNLI-VE       & $1.0{\times}10^{-5}$ & $3.0{\times}10^{-5}$ & $3.0{\times}10^{-5}$ & $3.0{\times}10^{-5}$ \\
IU-XRay      & $1.0{\times}10^{-5}$ & $3.0{\times}10^{-5}$ & $3.0{\times}10^{-5}$ & $3.0{\times}10^{-5}$ \\
\bottomrule
\end{tabular}
\end{table}

\begin{table}[h]
\caption{Target layers edited by Geodesic Unlearning. We use $\eta=0.75$ and $k=8$ for all configurations.}
\label{tab:app_gu_target_layers}
\centering
\begin{tabular}{lll}
\toprule
Model Family & Variant & Target Layer \\
\midrule
Fusion / Fusion (strong) & Concat & \texttt{fusion\_mlp0} \\
Fusion / Fusion (strong) & Gate & \texttt{gate\_mlp0} \\
Fusion / Fusion (strong) & Cross-Attention & \texttt{attn\_v\_from\_t\_in\_proj} \\
Two-tower & Late Concat & \texttt{fusion\_mlp0} \\
Two-tower & Mean Pooling & \texttt{text\_proj} \\
VLM & BLIP-2 Classifier Head & \texttt{qformer\_crossattn0\_out} \\
\bottomrule
\end{tabular}
\end{table}

\clearpage

\section{Supplementary experimental results}

\subsection{Performance stability analysis}
\label{app:performance_stability}

\subsubsection{Three-run means and standard deviations}
\label{app:performance_std}

Tables~\ref{tab:exp_main_performance_std} and~\ref{tab:snlive_robustness_std} supplement
Tables~\ref{tab:exp_main_performance} and~\ref{tab:snlive_robustness}, respectively,
with mean $\pm$ sample standard deviation over three independent runs. For the family-level results, we first average the variants within each seed and then compute
the mean and sample standard deviation across the three seed-level averages. The resulting standard deviations therefore describe variability across runs at the same
aggregation level as the main-text means.

The three-run results retain GU's highest mean TI-T F1 on all three datasets in
Table~\ref{tab:exp_main_performance_std}.
The family-level results in Table~\ref{tab:snlive_robustness_std} further show that GU
matches or improves upon interpolation's mean missing-modality recovery and full-modality
preservation across the evaluated families.

\begin{table}[t]
\caption{
Supplementary results for Table~\ref{tab:exp_main_performance}.
We report mean $\pm$ sample standard deviation over three independent runs.
\underline{Underline} and \textbf{bold} denote the second-best and best mean results, respectively.
}
\label{tab:exp_main_performance_std}
\centering

\begin{tabular}{lccc}
\toprule
Method & HM & IU-XRay & SNLI-VE \\
\midrule

Null-token
& $57.63\pm0.88$
& $80.00\pm0.47$
& $66.35\pm0.38$ \\

SMIL
& $56.60\pm0.74$
& $79.21\pm0.68$
& $\underline{72.20}\pm0.52$ \\

Flex-MoE
& $47.58\pm1.38$
& $71.02\pm0.82$
& $70.74\pm0.64$ \\

DyMo
& $48.82\pm0.93$
& $79.19\pm0.77$
& $61.30\pm0.70$ \\

Interpolation
& $\underline{58.50}\pm0.91$
& $\underline{80.14}\pm0.49$
& $67.29\pm0.61$ \\

\rowcolor{cyan!10}
\textbf{GU (Ours)}
& $\mathbf{60.88}\pm0.73$
& $\mathbf{81.41}\pm0.63$
& $\mathbf{74.41}\pm0.77$ \\

\bottomrule
\end{tabular}
\end{table}

\begin{table}[t]
\caption{
Supplementary robustness--preservation results for
Table~\ref{tab:snlive_robustness} on SNLI-VE.
Results are mean $\pm$ sample standard deviation over three independent runs.
}
\label{tab:snlive_robustness_std}
\centering

\begin{tabular}{llcccc}
\toprule
Metric & Method
& Fusion
& Fusion (str.)
& CLIP
& VLM \\
\midrule

TI-T
& Interp.
& $71.34\pm0.46$
& $72.34\pm0.38$
& $74.50\pm0.44$
& $70.65\pm0.79$ \\

\rowcolor{cyan!10}
& \textbf{GU}
& $\mathbf{73.86}\pm0.56$
& $\mathbf{73.06}\pm0.43$
& $\mathbf{74.68}\pm0.52$
& $\mathbf{70.65}\pm0.68$ \\

\midrule

TI-TI
& Interp.
& $86.33\pm0.18$
& $75.16\pm0.37$
& $76.13\pm0.38$
& $81.72\pm0.30$ \\

\rowcolor{cyan!10}
& \textbf{GU}
& $\mathbf{86.37}\pm0.14$
& $\mathbf{75.95}\pm0.31$
& $\mathbf{76.70}\pm0.31$
& $\mathbf{81.73}\pm0.18$ \\

\midrule

$\Delta$T
& Interp.
& $0.98\pm0.30$
& $0.96\pm0.22$
& $1.04\pm0.35$
& $0.01\pm0.10$ \\

\rowcolor{cyan!10}
& \textbf{GU}
& $\mathbf{3.49}\pm0.26$
& $\mathbf{1.69}\pm0.20$
& $\mathbf{1.22}\pm0.33$
& $\mathbf{0.01}\pm0.04$ \\

\midrule

$\Delta$TI
& Interp.
& $-0.13\pm0.11$
& $-1.13\pm0.31$
& $-1.21\pm0.34$
& $0.00\pm0.13$ \\

\rowcolor{cyan!10}
& \textbf{GU}
& $\mathbf{-0.09}\pm0.06$
& $\mathbf{-0.34}\pm0.24$
& $\mathbf{-0.64}\pm0.25$
& $\mathbf{+0.01}\pm0.04$ \\

\bottomrule
\end{tabular}
\end{table}

\subsubsection{Ten-seed paired significance tests}
\label{app:ten_seed_significance}

We further validate the gains in Table~\ref{tab:exp_main_performance} over ten matched seeds. For each dataset, GU is compared with the null-token baseline and the strongest competing method in that table: interpolation on HM and IU-XRay, and SMIL on SNLI-VE. We report the paired GU-minus-baseline F1 difference ($\Delta$F1) as mean $\pm$ sample standard deviation. Statistical significance is assessed using two-sided exact paired sign-flip tests, with Holm correction across all six comparisons.

As shown in Table~\ref{tab:significance_tests}, all six comparisons remain significant after Holm correction ($p<0.05$). Against the strongest competing method on each dataset, GU wins all 30 paired runs, with mean gains of $+2.41$, $+1.34$, and $+2.21$ F1 on HM, IU-XRay, and SNLI-VE, respectively. These results provide additional statistical support for the three-run main results reported separately above.

\begin{table}[t]
\caption{Ten-seed paired tests for GU against Null and the strongest competing method in Table~\ref{tab:exp_main_performance}. All six comparisons remain significant after Holm correction.}
\label{tab:significance_tests}
\centering

\begin{tabular}{llcccc}
\toprule
Dataset & Comparison & 10-seed paired $\Delta$F1 & GU wins & Raw $p$ & Holm $p$ \\
\midrule

HM
& GU$-$Null
& $+3.19\pm1.14$
& 9/10
& 0.003906
& 0.011719 \\

HM
& GU$-$Interp.
& $+2.41\pm0.77$
& 10/10
& 0.001953
& 0.011719 \\

IX
& GU$-$Null
& $+1.51\pm1.71$
& 9/10
& 0.033203
& 0.033203 \\

IX
& GU$-$Interp.
& $+1.34\pm0.45$
& 10/10
& 0.001953
& 0.011719 \\

SV
& GU$-$Null
& $+8.09\pm0.22$
& 10/10
& 0.001953
& 0.011719 \\

SV
& GU$-$SMIL
& $+2.21\pm0.68$
& 10/10
& 0.001953
& 0.011719 \\

\bottomrule
\end{tabular}
\end{table}

\subsection{More details on representation-level evidence}
\label{app:repr_evidence_details}

We quantify the representation similarity and functional changes behind
Figure~\ref{fig:exp_repr_ablation}(a) using the SNLI-VE Fusion Cross-Attention model.
Table~\ref{tab:fig4a_quantitative} reports mean $\pm$ sample standard deviation over
three independent runs using the same missing-image test inputs (TI-T).
Linear CKA measures the similarity between fusion-layer representations of each edited
model and the original multimodal model under the same TI-T condition.
Error-margin gain is measured relative to the original TI-T model.
The correction and regression rates denote the fractions of originally incorrect
predictions corrected and originally correct predictions broken after editing,
respectively.

Both edits preserve the broad representation structure, with linear CKA of $0.988$
for interpolation and $0.991$ for GU, consistent with their similar UMAP layouts in
Figure~\ref{fig:exp_repr_ablation}(a).
Despite this global similarity, GU produces substantially larger task-level changes:
its error-margin gain is $+0.30$ versus $+0.06$, and it corrects $34.0\%$ of the original
errors versus $11.8\%$ for interpolation, while increasing the regression rate by only
$0.6$ percentage points ($5.1\%$ versus $4.5\%$).
Correspondingly, TI-T F1 reaches $74.41$ with GU versus $67.29$ with interpolation.
These results show that GU preserves overall representation structure while making
targeted changes that improve task-relevant margins and correct substantially more errors.

\begin{table}[t]
\caption{
Quantitative analysis of Figure~\ref{fig:exp_repr_ablation}(a) on SNLI-VE Fusion Cross-Attention.
Both edits preserve high representation similarity, while GU yields larger task-level improvements.
}
\label{tab:fig4a_quantitative}
\centering

\begin{tabular}{lcccc}
\toprule
Method
& TI-T
& CKA to orig.
& Error-margin gain
& Corr./regr. rate (\%) \\
\midrule

Orig.\ TI-T
& $66.35\pm0.38$
& $1.000$
& $0$
& $0/0$ \\

Interp.
& $67.29\pm0.61$
& $0.988\pm0.001$
& $+0.06\pm0.01$
& $11.8\pm0.6 \,/\, 4.5\pm0.2$ \\

\rowcolor{cyan!10}
\textbf{GU (Ours)}
& $74.41\pm0.77$
& $0.991\pm0.001$
& $+0.30\pm0.02$
& $34.0\pm1.3 \,/\, 5.1\pm0.3$ \\

\bottomrule
\end{tabular}
\end{table}

\subsection{More results on robustness--preservation trade-off}
\label{app:preservation_tradeoff}

The main text reports family-level averages for robustness recovery and full-modality preservation on SNLI-VE. Table~\ref{tab:variant_preservation_snlive} provides complete per-variant results, reported as mean $\pm$ sample standard deviation over three independent runs. For each model variant, TI-T and TI-TI denote missing-modality and full-modality performance after editing, while $\Delta T$ and $\Delta TI$ measure their changes relative to the original TI model.

Across variants, GU generally yields larger TI-T gains than layer interpolation while producing smaller or comparable TI-TI changes. The difference is most pronounced in cross-attention fusion, where GU substantially improves missing-modality robustness with limited full-modality degradation. For CLIP-style and BLIP-2 architectures, the gains are smaller, consistent with their more distributed cross-modal interactions. Overall, the per-variant results reinforce the family-level robustness--preservation pattern reported in Section~\ref{sec:exp_preservation}.

\begin{table*}[t]
\caption{
Per-variant robustness--preservation results on SNLI-VE. GU generally provides stronger recovery with comparable or better full-modality preservation than interpolation.
}
\label{tab:variant_preservation_snlive}
\centering
\small
\setlength{\tabcolsep}{4pt}

\begin{tabular}{
@{}
p{1.05cm}
l
l
l
c
c
c
c
@{}
}
\toprule
Family & Variant & Target Layer & Method
& TI-T & TI-TI & $\Delta T$ & $\Delta TI$ \\
\midrule

\multirow{6}{*}{Fusion}
& Concat
& fusion\_mlp0
& Interp.
& $74.86\pm.68$
& $86.10\pm.16$
& $1.82\pm.48$
& $-0.12\pm.10$ \\

& \cellcolor{cyan!10}Concat
& \cellcolor{cyan!10}fusion\_mlp0
& \cellcolor{cyan!10}\textbf{GU}
& \cellcolor{cyan!10}$\mathbf{75.11}\pm.56$
& \cellcolor{cyan!10}$\mathbf{86.14}\pm.12$
& \cellcolor{cyan!10}$\mathbf{2.07}\pm.32$
& \multicolumn{1}{>{\columncolor{cyan!10}[\tabcolsep][0pt]}c@{}}{$\mathbf{-0.08}\pm.05$} \\

& Gate
& gate\_mlp0
& Interp.
& $71.88\pm.47$
& $86.40\pm.18$
& $0.18\pm.24$
& $-0.13\pm.11$ \\

& \cellcolor{cyan!10}Gate
& \cellcolor{cyan!10}gate\_mlp0
& \cellcolor{cyan!10}\textbf{GU}
& \cellcolor{cyan!10}$\mathbf{72.05}\pm.35$
& \cellcolor{cyan!10}$\mathbf{86.44}\pm.12$
& \cellcolor{cyan!10}$\mathbf{0.35}\pm.07$
& \multicolumn{1}{>{\columncolor{cyan!10}[\tabcolsep][0pt]}c@{}}{$\mathbf{-0.09}\pm.04$} \\

& Cross-Attn
& attn\_v$\leftarrow$t\_in
& Interp.
& $67.29\pm.61$
& $86.50\pm.24$
& $0.94\pm.29$
& $-0.14\pm.15$ \\

& \cellcolor{cyan!10}Cross-Attn
& \cellcolor{cyan!10}attn\_v$\leftarrow$t\_in
& \cellcolor{cyan!10}\textbf{GU}
& \cellcolor{cyan!10}$\mathbf{74.41}\pm.77$
& \cellcolor{cyan!10}$\mathbf{86.53}\pm.17$
& \cellcolor{cyan!10}$\mathbf{8.06}\pm.40$
& \multicolumn{1}{>{\columncolor{cyan!10}[\tabcolsep][0pt]}c@{}}{$\mathbf{-0.11}\pm.08$} \\

\midrule

\multirow{6}{*}{\shortstack[l]{Fusion\\(strong)}}
& Concat
& fusion\_mlp0
& Interp.
& $72.93\pm.33$
& $76.62\pm.18$
& $0.15\pm.16$
& $0.18\pm.13$ \\

& \cellcolor{cyan!10}Concat
& \cellcolor{cyan!10}fusion\_mlp0
& \cellcolor{cyan!10}\textbf{GU}
& \cellcolor{cyan!10}$\mathbf{73.06}\pm.24$
& \cellcolor{cyan!10}$\mathbf{76.74}\pm.12$
& \cellcolor{cyan!10}$\mathbf{0.28}\pm.05$
& \multicolumn{1}{>{\columncolor{cyan!10}[\tabcolsep][0pt]}c@{}}{$\mathbf{0.30}\pm.07$} \\

& Gate
& gate\_mlp0
& Interp.
& $70.13\pm.52$
& $73.09\pm.23$
& $2.48\pm.34$
& $0.37\pm.17$ \\

& \cellcolor{cyan!10}Gate
& \cellcolor{cyan!10}gate\_mlp0
& \cellcolor{cyan!10}\textbf{GU}
& \cellcolor{cyan!10}$\mathbf{71.99}\pm.70$
& \cellcolor{cyan!10}$\mathbf{73.30}\pm.16$
& \cellcolor{cyan!10}$\mathbf{4.34}\pm.46$
& \multicolumn{1}{>{\columncolor{cyan!10}[\tabcolsep][0pt]}c@{}}{$\mathbf{0.58}\pm.09$} \\

& Cross-Attn
& attn\_v$\leftarrow$t\_in
& Interp.
& $73.95\pm.48$
& $75.76\pm.82$
& $0.26\pm.24$
& $-3.95\pm.75$ \\

& \cellcolor{cyan!10}Cross-Attn
& \cellcolor{cyan!10}attn\_v$\leftarrow$t\_in
& \cellcolor{cyan!10}\textbf{GU}
& \cellcolor{cyan!10}$\mathbf{74.13}\pm.35$
& \cellcolor{cyan!10}$\mathbf{77.81}\pm.65$
& \cellcolor{cyan!10}$\mathbf{0.44}\pm.09$
& \multicolumn{1}{>{\columncolor{cyan!10}[\tabcolsep][0pt]}c@{}}{$\mathbf{-1.90}\pm.57$} \\

\midrule

\multirow{4}{*}[-0.8\baselineskip]{\shortstack[l]{Two-\\tower}}
& \shortstack[l]{Late\\Concat}
& fusion\_mlp0
& Interp.
& $73.72\pm.42$
& $77.61\pm.22$
& $0.95\pm.24$
& $-0.44\pm.17$ \\

& \cellcolor{cyan!10}\shortstack[l]{Late\\Concat}
& \cellcolor{cyan!10}fusion\_mlp0
& \cellcolor{cyan!10}\textbf{GU}
& \cellcolor{cyan!10}$\mathbf{73.78}\pm.51$
& \cellcolor{cyan!10}$\mathbf{77.74}\pm.15$
& \cellcolor{cyan!10}$\mathbf{1.01}\pm.30$
& \multicolumn{1}{>{\columncolor{cyan!10}[\tabcolsep][0pt]}c@{}}{$\mathbf{-0.31}\pm.09$} \\

& \shortstack[l]{Mean\\Pooling}
& text\_proj1
& Interp.
& $75.28\pm.67$
& $74.65\pm.59$
& $1.13\pm.52$
& $-1.97\pm.55$ \\

& \cellcolor{cyan!10}\shortstack[l]{Mean\\Pooling}
& \cellcolor{cyan!10}text\_proj1
& \cellcolor{cyan!10}\textbf{GU}
& \cellcolor{cyan!10}$\mathbf{75.58}\pm.53$
& \cellcolor{cyan!10}$\mathbf{75.65}\pm.46$
& \cellcolor{cyan!10}$\mathbf{1.43}\pm.35$
& \multicolumn{1}{>{\columncolor{cyan!10}[\tabcolsep][0pt]}c@{}}{$\mathbf{-0.97}\pm.41$} \\

\midrule

\multirow{2}{*}{VLM}
& BLIP-2
& q\_crossattn0\_o
& Interp.
& $70.65\pm.79$
& $81.72\pm.30$
& $0.01\pm.10$
& $0.00\pm.13$ \\

& \cellcolor{cyan!10}BLIP-2
& \cellcolor{cyan!10}q\_crossattn0\_o
& \cellcolor{cyan!10}\textbf{GU}
& \cellcolor{cyan!10}$\mathbf{70.65}\pm.68$
& \cellcolor{cyan!10}$\mathbf{81.73}\pm.18$
& \cellcolor{cyan!10}$\mathbf{0.01}\pm.04$
& \multicolumn{1}{>{\columncolor{cyan!10}[\tabcolsep][0pt]}c@{}}{$\mathbf{0.01}\pm.04$} \\

\bottomrule
\end{tabular}
\end{table*}

\subsection{Comparisons with more baselines}
\label{app:additional_baselines}

\subsubsection{MoRA and OGM-GE}
\label{app:mora_ogm}

\begin{table*}[t]
\caption{
Cross-architecture comparison with MoRA and OGM-GE. GU achieves the highest mean TI-T in all nine evaluated settings.
Entries report TI-T / TI-TI F1.
$\Delta$TI denotes GU's mean TI-TI change relative to Null-token
(GU minus Null-token).
}
\label{tab:mora_ogm_cross_arch}
\centering
\small
\setlength{\tabcolsep}{4pt}

\begin{tabular}{lccc}
\toprule
Method
& \shortstack{Fusion\\Concat}
& \shortstack{CLIP Late\\Concat}
& \shortstack{CLIP Mean\\Pooling} \\
\midrule

\multicolumn{4}{l}{\textit{Hateful Memes}} \\

Null-token
& $68.96\pm0.44 \,/\, 69.46\pm0.32$
& $66.58\pm0.41 \,/\, 67.31\pm0.27$
& $66.87\pm0.24 \,/\, 67.14\pm0.20$ \\

MoRA
& -- \,/\, --
& $65.64\pm0.28 \,/\, 65.39\pm0.83$
& $60.45\pm6.17 \,/\, 64.98\pm0.87$ \\

OGM-GE
& $66.71\pm0.56 \,/\, 67.07\pm0.78$
& $66.07\pm0.47 \,/\, 65.85\pm0.41$
& $64.51\pm0.66 \,/\, 65.68\pm1.10$ \\

\rowcolor{cyan!10}
\textbf{GU}
& $69.01\pm0.24 \,/\, 69.32\pm0.30$
& $67.03\pm0.17 \,/\, 67.26\pm0.31$
& $66.96\pm0.34 \,/\, 67.09\pm0.22$ \\

GU $\Delta$TI
& $-0.14$ & $-0.05$ & $-0.05$ \\

\midrule

\multicolumn{4}{l}{\textit{IU-XRay}} \\

Null-token
& $80.70\pm0.38 \,/\, 86.41\pm0.22$
& $79.45\pm0.29 \,/\, 80.85\pm0.19$
& $74.30\pm0.38 \,/\, 78.00\pm0.20$ \\

MoRA
& -- \,/\, --
& $73.87\pm1.91 \,/\, 78.14\pm0.26$
& $70.55\pm2.06 \,/\, 75.13\pm0.45$ \\

OGM-GE
& $72.62\pm3.87 \,/\, 77.53\pm2.25$
& $75.04\pm0.48 \,/\, 75.51\pm1.02$
& $67.54\pm0.89 \,/\, 71.36\pm0.33$ \\

\rowcolor{cyan!10}
\textbf{GU}
& $81.69\pm0.72 \,/\, 85.33\pm0.80$
& $80.16\pm0.24 \,/\, 80.57\pm0.31$
& $75.98\pm0.61 \,/\, 77.28\pm0.36$ \\

GU $\Delta$TI
& $-1.08$ & $-0.28$ & $-0.72$ \\

\midrule

\multicolumn{4}{l}{\textit{SNLI-VE}} \\

Null-token
& $73.04\pm0.24 \,/\, 86.22\pm0.07$
& $72.77\pm0.21 \,/\, 78.05\pm0.06$
& $74.15\pm0.18 \,/\, 76.62\pm0.06$ \\

MoRA
& -- \,/\, --
& $70.28\pm0.99 \,/\, 75.06\pm0.19$
& $72.21\pm0.12 \,/\, 72.17\pm0.04$ \\

OGM-GE
& $67.26\pm0.33 \,/\, 83.91\pm0.24$
& $68.10\pm2.14 \,/\, 71.07\pm0.10$
& $68.00\pm0.54 \,/\, 70.30\pm0.21$ \\

\rowcolor{cyan!10}
\textbf{GU}
& $75.11\pm0.56 \,/\, 86.14\pm0.12$
& $73.78\pm0.51 \,/\, 77.74\pm0.15$
& $75.58\pm0.53 \,/\, 75.65\pm0.46$ \\

GU $\Delta$TI
& $-0.08$ & $-0.31$ & $-0.97$ \\

\bottomrule
\end{tabular}
\end{table*}

We further compare GU with two architecture-compatible baselines. MoRA~\cite{zhao2026mora} is a parameter-efficient missing-modality adaptation method and is evaluated on CLIP Late Concat and Mean Pooling. Its formulation couples modality-specific and shared low-rank updates across paired text and vision encoder weights. Applying it to Fusion Concat would require new insertion points and a fusion-specific sharing rule, so we retain its matched dual-encoder parameterization.

OGM-GE~\cite{peng2022balanced} is a training-time modality-balancing method and is evaluated on Fusion Concat and both CLIP variants. CLIP Mean Pooling permits exact decomposition of the full logit into modality contributions. For nonlinear Late Concat, we retain the published discrepancy, gradient-modulation, and generalization-enhancement rules while estimating modality confidence through native unimodal forward paths.

Table~\ref{tab:mora_ogm_cross_arch} reports TI-T / TI-TI as mean $\pm$ sample standard deviation over three independent runs. The $\Delta$TI rows denote GU's mean TI-TI change relative to Null-token. GU achieves the highest mean TI-T in all nine settings, including all six CLIP settings. Across the six CLIP comparisons, its average advantages over MoRA and OGM-GE are $4.42$ and $5.04$ F1, respectively. Recovery over Null-token is larger on IU-XRay and SNLI-VE. On HM Fusion Concat, the original missing-modality gap is only $0.50$ F1, leaving little room for further recovery.

Across all nine settings, GU improves TI-T by $0.94$ F1 on average while changing TI-TI by only $-0.41$ F1; the largest TI-TI loss is $1.08$ F1 on IU-XRay Fusion Concat. MoRA and OGM-GE do not consistently improve missing-image performance, whereas GU remains effective across both CLIP variants beyond the original fusion-only comparison.

\subsubsection{LoRA and Dual CE}
\label{app:dual_ce}

\begin{table}[t]
\caption{
Comparison with post-hoc LoRA and Dual CE on SNLI-VE.
Entries report F1 as mean $\pm$ sample standard deviation over three independent runs.
}
\label{tab:dual_ce_comparison}
\centering

\begin{tabular}{llcc}
\toprule
Architecture & Method & TI-T & TI-TI \\
\midrule

\multirow{5}{*}{Fusion CA}
& Null-token
& $66.35\pm0.38$
& $86.64\pm0.10$ \\

& LoRA
& $72.37\pm0.25$
& $85.02\pm0.35$ \\

& Dual CE
& $79.92\pm0.23$
& $85.70\pm0.22$ \\

& \cellcolor{cyan!10}\textbf{GU}
& \cellcolor{cyan!10}$74.41\pm0.77$
& \cellcolor{cyan!10}$86.53\pm0.17$ \\

& \cellcolor{cyan!10}Dual CE+GU
& \cellcolor{cyan!10}$80.09\pm0.30$
& \cellcolor{cyan!10}$86.51\pm0.19$ \\

\midrule

\multirow{5}{*}{CLIP MP}
& Null-token
& $74.15\pm0.18$
& $76.62\pm0.06$ \\

& LoRA
& $75.18\pm0.07$
& $72.96\pm0.64$ \\

& Dual CE
& $76.31\pm0.13$
& $75.22\pm0.08$ \\

& \cellcolor{cyan!10}\textbf{GU}
& \cellcolor{cyan!10}$75.58\pm0.53$
& \cellcolor{cyan!10}$75.65\pm0.46$ \\

& \cellcolor{cyan!10}Dual CE+GU
& \cellcolor{cyan!10}$75.46\pm0.61$
& \cellcolor{cyan!10}$75.45\pm0.78$ \\

\midrule

\multirow{5}{*}{BLIP-2}
& Null-token
& $70.64\pm0.70$
& $81.72\pm0.18$ \\

& LoRA
& $70.90\pm0.86$
& $81.04\pm0.23$ \\

& Dual CE
& $75.05\pm0.13$
& $81.19\pm0.16$ \\

& \cellcolor{cyan!10}\textbf{GU}
& \cellcolor{cyan!10}$70.65\pm0.68$
& \cellcolor{cyan!10}$81.73\pm0.18$ \\

& \cellcolor{cyan!10}Dual CE+GU
& \cellcolor{cyan!10}$75.05\pm0.17$
& \cellcolor{cyan!10}$81.04\pm0.49$ \\

\bottomrule
\end{tabular}
\end{table}

We compare GU with post-hoc LoRA and Dual CE on SNLI-VE using
Fusion Cross-Attention (Fusion CA), CLIP Mean Pooling (CLIP MP),
and BLIP-2. Table~\ref{tab:dual_ce_comparison} reports performance
under missing-image (TI-T) and full-modality (TI-TI) evaluation.

\paragraph{Comparison with LoRA.}
LoRA starts from the same trained TI checkpoint as GU and adapts
the same target layer with rank $8$. It trains on missing-image
inputs for $4$ epochs at learning rate $10^{-4}$, selecting the
best checkpoint on the development set. GU instead uses the
matched unimodal reference for a single closed-form edit,
without further gradient updates to the TI checkpoint.

GU achieves higher mean TI-T and TI-TI than LoRA on Fusion CA
and CLIP MP. On BLIP-2, LoRA obtains slightly greater TI-T
recovery but reduces TI-TI, whereas GU leaves both metrics
essentially unchanged.
Averaged across the three architectures, LoRA improves TI-T
by $2.44$ F1 with a TI-TI change of $-1.99$ F1 relative to
Null-token; GU improves TI-T by $3.17$ F1 with a TI-TI change
of only $-0.36$ F1. Thus, GU provides $0.73$ F1 greater
missing-image recovery with $1.63$ F1 less full-modality
degradation on average. These results demonstrate a stronger
average cross-architecture robustness--preservation trade-off
than matched-layer LoRA adaptation, extending the comparison
beyond layer interpolation and the fusion setting.

\paragraph{Comparison with Dual CE.}
Dual CE retrains each architecture from scratch with a shared
classifier jointly supervising full-input and missing-image
paths, using an auxiliary weight of $0.5$. Dual CE+GU applies
GU to the resulting Dual CE checkpoint. Dual CE is a strong retraining baseline, improving TI-T over
Null-token by $13.57$, $2.16$, and $4.41$ F1 on Fusion CA,
CLIP MP, and BLIP-2, respectively. This recovery comes with
TI-TI changes of $-0.94$, $-1.40$, and $-0.53$ F1, whereas
GU preserves higher full-modality performance on all three
architectures.

The two approaches are complementary in the Fusion setting:
applying GU after Dual CE raises TI-TI from $85.70$ to $86.51$
while slightly improving TI-T from $79.92$ to $80.09$.
On CLIP and BLIP-2, the composition provides no additional
recovery benefit. Overall, Dual CE favors stronger missing-image
recovery through retraining, while GU provides a post-training
correction with stronger full-modality preservation.
The computational costs of GU, LoRA, and Dual CE are compared
in Table~\ref{tab:efficiency_cost}.

\subsection{Controlled interventions for geometric correction}
\label{app:controlled_interventions}

We examine two design choices underlying GU: \emph{where to edit and how to edit}. These controlled comparisons connect the diagnostics in Section~\ref{sec:analysis} with the geometric rationale in Appendix~\ref{app:gu_geometric_rationale}.

\subsubsection{Target-layer selection}
\label{app:layer_selection_controls}

\begin{table*}[t]
\caption{
Target-layer selection on IU-XRay Fusion Cross-Attention. Joint angle--harm selection yields the strongest recovery and preservation.
}
\label{tab:layer_selection_ablation}
\centering

\begin{tabular}{lcc>{\columncolor{cyan!10}}c}
\toprule
& Angle only & Harm only & \textbf{Angle+harm} \\
\midrule

Selected layer
& \texttt{attn\_t\_from\_v\_out\_proj}
& \texttt{text\_proj}
& \texttt{attn\_v\_from\_t\_in\_proj} \\

TI-T
& $80.34\pm0.66$
& $80.72\pm0.40$
& $\mathbf{81.41}\pm0.63$ \\

TI-TI
& $83.96\pm0.37$
& $84.21\pm0.32$
& $\mathbf{84.44}\pm0.28$ \\

$\Delta T$
& $+0.34$
& $+0.72$
& $\mathbf{+1.41}$ \\

$\Delta TI$
& $-0.61$
& $-0.36$
& $\mathbf{-0.13}$ \\

\bottomrule
\end{tabular}
\end{table*}

Using the IU-XRay Fusion Cross-Attention layer sweep in Figure~\ref{fig:exp_repr_ablation}(b), we compare angle-only, harm-only, and joint angle--harm selection while keeping checkpoints and the editing protocol fixed. Angle-only and harm-only rank candidate layers by principal-subspace deviation and task-aware harm, respectively, while the joint rule uses the development-set score in Eq.~\eqref{eq:target_layer_score}. Table~\ref{tab:layer_selection_ablation} reports mean $\pm$ sample standard deviation over three independent runs. $\Delta T$ and $\Delta TI$ denote changes in TI-T and TI-TI relative to the original TI model.

The joint rule produces the largest TI-T gain ($+1.41$ F1) and smallest TI-TI loss ($-0.13$ F1). Angle-only and harm-only selection yield smaller gains of $+0.34$ and $+0.72$, with TI-TI changes of $-0.61$ and $-0.36$, respectively. Combining geometric deviation with task-aware harm therefore identifies a more effective correction target than either signal alone.

\subsubsection{Editing strategy comparison}
\label{app:editing_strategy_controls}

\begin{table}[t]
\caption{
Controlled comparison of editing strategies on Fusion Cross-Attention. GU achieves the highest mean TI-T and TI-TI across all three datasets.
Entries report TI-T / TI-TI F1.
}
\label{tab:controlled_interventions}
\centering
\small
\setlength{\tabcolsep}{4pt}

\begin{tabular}{lcc>{\columncolor{cyan!10}}c}
\toprule
Dataset
& Weight
& Linear
& \textbf{GU (Ours)} \\
\midrule

HM
& $58.50\pm0.91 \,/\, 66.52\pm0.33$
& $58.07\pm1.03 \,/\, 66.62\pm0.21$
& $\mathbf{60.88}\pm0.73 \,/\, \mathbf{67.55}\pm0.26$ \\

IX
& $80.14\pm0.49 \,/\, 82.67\pm0.56$
& $79.83\pm0.75 \,/\, 82.39\pm0.36$
& $\mathbf{81.41}\pm0.63 \,/\, \mathbf{84.44}\pm0.28$ \\

SV
& $67.29\pm0.61 \,/\, 86.50\pm0.24$
& $66.84\pm1.29 \,/\, 86.50\pm0.07$
& $\mathbf{74.41}\pm0.77 \,/\, \mathbf{86.53}\pm0.17$ \\

\bottomrule
\end{tabular}
\end{table}

We next compare three editing strategies on Fusion Cross-Attention across the three datasets. All methods use matched TI/T checkpoints, the same target layer and reference, and the same evaluation protocol. \textbf{Weight} interpolates the full selected weight matrix with its reference using fixed $\alpha=0.5$, thereby changing input orientation together with output factors, scales, and residual. \textbf{Linear} and \textbf{GU} instead edit the aligned principal input subspace while retaining the original TI output factor, principal coefficients, and residual. Linear applies polar-retracted interpolation as in Eq.~\eqref{eq:app_gu_linear_polar}, whereas GU follows the Grassmannian geodesic. Both use $k=8$ and $\eta=0.75$. Table~\ref{tab:controlled_interventions} reports TI-T / TI-TI F1 as mean $\pm$ sample standard deviation over three independent runs.

GU achieves the highest mean TI-T and TI-TI in all three settings. On HM and IU-XRay, it improves both missing-image recovery and full-modality preservation over Weight and Linear. On SNLI-VE, GU reaches $74.41$ TI-T F1 versus $67.29$ for Weight and $66.84$ for Linear, while maintaining comparable TI-TI performance. In particular, the comparison with Linear isolates the interpolation path: even with the same target subspace, rank, coefficient, and retained components, the geodesic yields substantially stronger recovery.

Together, these controls support the two key design choices in GU: combining geometric deviation with task-aware harm for target selection, and correcting the selected principal input subspace along a Grassmannian geodesic.

\subsection{Generalization to multiple-choice VQA}
\label{app:aokvqa}

\begin{table}[t]
\caption{
A-OKVQA multiple-choice accuracy (\%) over ten runs. GU provides the strongest full-modality preservation among the evaluated interventions.
Bold marks the best result among Interpolation, LoRA, and GU.
}
\label{tab:aokvqa_10seed}
\centering

\begin{tabular}{lcc}
\toprule
Method & TI-T Acc. & TI-TI Acc. \\
\midrule

T Refer.\ (T-T)
& $47.17\pm0.31$
& -- \\

Shuffle-I
& --
& $39.85\pm1.97$ \\

Orig.\ TI
& $38.94\pm1.92$
& $50.79\pm0.71$ \\

Interp.
& $39.80\pm1.21$
& $43.21\pm1.44$ \\

LoRA
& $\mathbf{43.52}\pm0.60$
& $45.01\pm1.72$ \\

\rowcolor{cyan!10}
\textbf{GU (Ours)}
& $41.07\pm1.79$
& $\mathbf{47.40}\pm1.19$ \\

\bottomrule
\end{tabular}
\end{table}

To extend the evaluation beyond binary classification, we evaluate GU on the A-OKVQA~\cite{schwenk2022okvqa} multiple-choice VQA benchmark. We use the strong-encoder cross-attention fusion configuration described in Appendix~\ref{app:backbone_configs}, with Qwen2.5-3B and OpenCLIP ViT-bigG-14 encoders, and follow the official dataset splits.

Table~\ref{tab:aokvqa_10seed} reports accuracy as mean $\pm$ sample standard deviation over ten independent runs. TI-T and TI-TI denote missing-image and full-modality evaluation, respectively; the T reference reports T-T accuracy in the TI-T column. Shuffle-I evaluates the original TI model with shuffled images under full-input inference. Bold marks the best result among Interpolation, LoRA, and GU.

The ordering TI-TI $>$ T-T $>$ TI-T persists on A-OKVQA ($50.79 > 47.17 > 38.94$). Shuffling images reduces full-input accuracy from $50.79\%$ to $39.85\%$, confirming that image information contributes substantially to the full-input prediction.

GU improves TI-T accuracy by $2.13$ points and achieves the highest TI-TI accuracy among the three interventions, with a $3.39$-point loss relative to the original TI model. Its TI-TI accuracy exceeds Interpolation and LoRA by $4.19$ and $2.39$ points, respectively. LoRA achieves larger TI-T recovery ($+4.58$), but with a larger TI-TI loss ($-5.78$). These results extend GU's recovery--preservation behavior beyond binary classification to multiple-choice VQA.

\subsection{Complementary evaluation under missing-text inputs}
\label{app:missing_text}

\begin{table}[t]
\caption{
Missing-text evaluation with Fusion Cross-Attention. GU improves TI-I while largely preserving full-modality performance across all three datasets.
}
\label{tab:missing_text}
\centering

\begin{tabular}{llccc}
\toprule
Method
& Eval.
& HM
& IU-XRay
& SNLI-VE \\
\midrule

Orig.\ TI
& TI-TI
& $68.82\pm0.48$
& $84.57\pm0.28$
& $86.64\pm0.10$ \\

I Refer.
& I-I
& $66.58\pm0.15$
& $75.32\pm0.41$
& $68.74\pm0.00$ \\

Orig.\ TI
& TI-I
& $63.32\pm0.88$
& $73.58\pm0.85$
& $64.48\pm0.97$ \\

\midrule

Interp.
& TI-I
& $64.10\pm0.24$
& $73.70\pm0.76$
& $65.87\pm0.46$ \\

Interp.
& TI-TI
& $68.30\pm0.28$
& $84.39\pm0.36$
& $86.01\pm0.15$ \\

\midrule

\rowcolor{cyan!10}
\textbf{GU (Ours)}
& TI-I
& $64.34\pm0.20$
& $74.42\pm0.22$
& $66.61\pm0.27$ \\

\rowcolor{cyan!10}
\textbf{GU (Ours)}
& TI-TI
& $68.65\pm0.23$
& $84.64\pm0.48$
& $86.67\pm0.15$ \\

\bottomrule
\end{tabular}
\end{table}

We further evaluate GU under missing-text inputs using Fusion Cross-Attention on HM, IU-XRay, and SNLI-VE. Here, TI-I denotes a TI-trained model evaluated without text, while I-I denotes the matched image-only reference. Both GU and layer interpolation use this image-only model as the reference. Table~\ref{tab:missing_text} reports F1 as mean $\pm$ sample standard deviation over three independent runs.

The ordering TI-TI $>$ I-I $>$ TI-I holds on all three datasets, mirroring the missing-image pattern in Section~\ref{sec:analysis_missing_failure}. GU improves TI-I by $1.02$, $0.84$, and $2.13$ F1 on HM, IU-XRay, and SNLI-VE, respectively, while changing TI-TI by only $-0.17$, $+0.07$, and $+0.03$ F1.

GU also achieves higher mean TI-I than matched layer interpolation and remains closer to the original TI-TI performance on all three datasets. Although the edited TI-I remains below the matched I-I reference, GU consistently narrows this gap. These results demonstrate GU's generalizability across missingness directions, as the same correction principle consistently improves robustness for both missing-image and missing-text inputs when paired with the modality available at deployment.


\end{document}